\PassOptionsToPackage{prologue,dvipsnames}{xcolor}
\documentclass[journal]{vgtc}                     

\onlineid{1465}

\vgtccategory{Research}

\vgtcpapertype{please specify}

\title{ESIQA: Perceptual Quality Assessment of Vision-Pro-based Egocentric Spatial Images}

\author{%
Xilei Zhu, Liu Yang, Huiyu Duan*, Xiongkuo Min*,~\textit{Member,~IEEE}, Guangtao Zhai*,~\textit{Fellow,~IEEE}, \\
and Patrick Le Callet,~\textit{Fellow,~IEEE}
}

\authorfooter{
  \item
  	Xilei Zhu, Liu Yang, Huiyu Duan, Xiongkuo Min, and Guangtao Zhai are with the Institute of Image Communication and Network Engineering, Shanghai Jiao Tong University, China.
  	E-mail: \{xilei\_zhu | ylyl.yl | huiyuduan | minxiongkuo | zhaiguangtao\}@sjtu.edu.cn.
  \item
  	Patrick Le Callet is with the Institut Universitaire de France (IUF), University of Nantes, France.
  	E-mail: patrick.lecallet@univ-nantes.fr.
 \item
  	*Huiyu Duan, Xiongkuo Min, and Guangtao Zhai are co-corresponding authors.
}

\abstract{%
With the development of eXtended Reality (XR), photo capturing and display technology based on head-mounted displays (HMDs) have experienced significant advancements and gained considerable attention.
Egocentric spatial images and videos are emerging as a compelling form of stereoscopic XR content. 
The assessment for the Quality of Experience (QoE) of XR content is important to ensure a high-quality viewing experience.
Different from traditional 2D images, egocentric spatial images present challenges for perceptual quality assessment due to their special shooting, processing methods, and stereoscopic characteristics. However, the corresponding image quality assessment (IQA) research for egocentric spatial images is still lacking.
In this paper, we establish the \underline{\textbf{E}}gocentric \underline{\textbf{S}}patial \underline{\textbf{I}}mages \underline{\textbf{Q}}uality \underline{\textbf{A}}ssessment \underline{\textbf{D}}atabase (ESIQAD), the first IQA database dedicated for egocentric spatial images as far as we know. Our ESIQAD includes 500 egocentric spatial images and the corresponding mean opinion scores (MOSs) under three display modes, including 2D display, 3D-window display, and 3D-immersive display. 
Based on our ESIQAD, we propose a novel mamba2-based multi-stage feature fusion model, termed ESIQAnet, which predicts the perceptual quality of egocentric spatial images under the three display modes. 
Specifically, we first extract features from multiple visual state space duality (VSSD) blocks, then apply cross attention to fuse binocular view information and use transposed attention to further refine the features.
The multi-stage features are finally concatenated and fed into a quality regression network to predict the quality score.
Extensive experimental results demonstrate that the  ESIQAnet outperforms 22 state-of-the-art IQA models on the ESIQAD under all three display modes.
The database and code are available at \textcolor{magenta}{https://github.com/IntMeGroup/ESIQA}.
}


\teaser{
  \centering
  \includegraphics[width=0.755\linewidth, alt={A view of a city with buildings peeking out of the clouds.}]{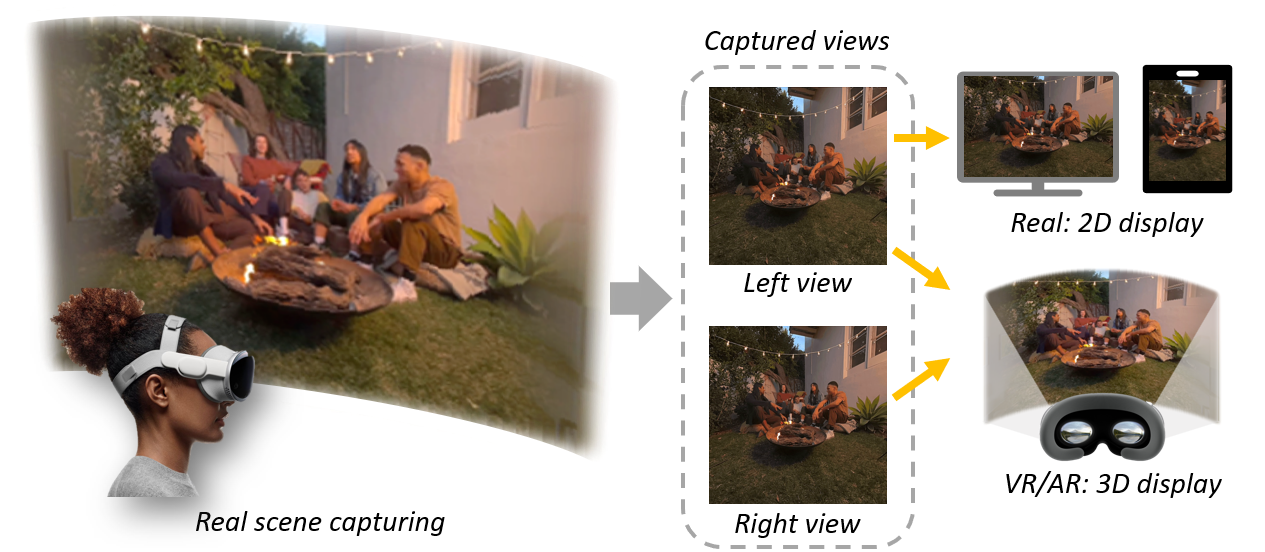}
  \caption{%
  	Illustration of the capturing and displaying methods for egocentric spatial images. Captured images, including left and right views, enable standard 2D views on conventional screens and immersive 3D experiences in VR/AR displays. 
  }
  \label{fig:display}
}

\graphicspath{{figs/}{figures/}{pictures/}{images/}{./}} 

\usepackage{booktabs}                  
\usepackage{mathptmx}                  
\usepackage{gensymb}
\usepackage{amsmath}
\usepackage{amsfonts} 
\usepackage{colortbl}
\definecolor{mygray}{gray}{0.91}

\begin{document}
\firstsection{Introduction}
\maketitle
\label{sec:intro}
Egocentric shooting and displaying technology has gained considerable attention recently, which can provide convenient capturing methods and immersive experiences through head-mounted displays (HMDs) \cite{duan2022confusing,duan2022saliency, duan2024quick}, such as Apple Vision Pro, Meta Quest 3, and Google Glass, \textit{etc}. 
Egocentric spatial images, captured from the first-person perspective, are emerging as an important form of stereoscopic content.
As shown in \cref{fig:display}, egocentric spatial images are captured with HMDs in real-world scenes and can be transmitted and viewed on various terminal devices. 
For binocular displays, since a spatial image is composed of a pair of disparity views to simulate human binocular vision, it is possible to directly create a three-dimensional (3D) effect and provide an immersive experience in HMDs.
For conventional flat displays, generally only the left view, \textit{i.e.,} a standard two-dimensional (2D) image, is presented.

\begin{figure*}[t]
\vspace{-15pt}
    \centering
    \includegraphics[width=0.95\linewidth]{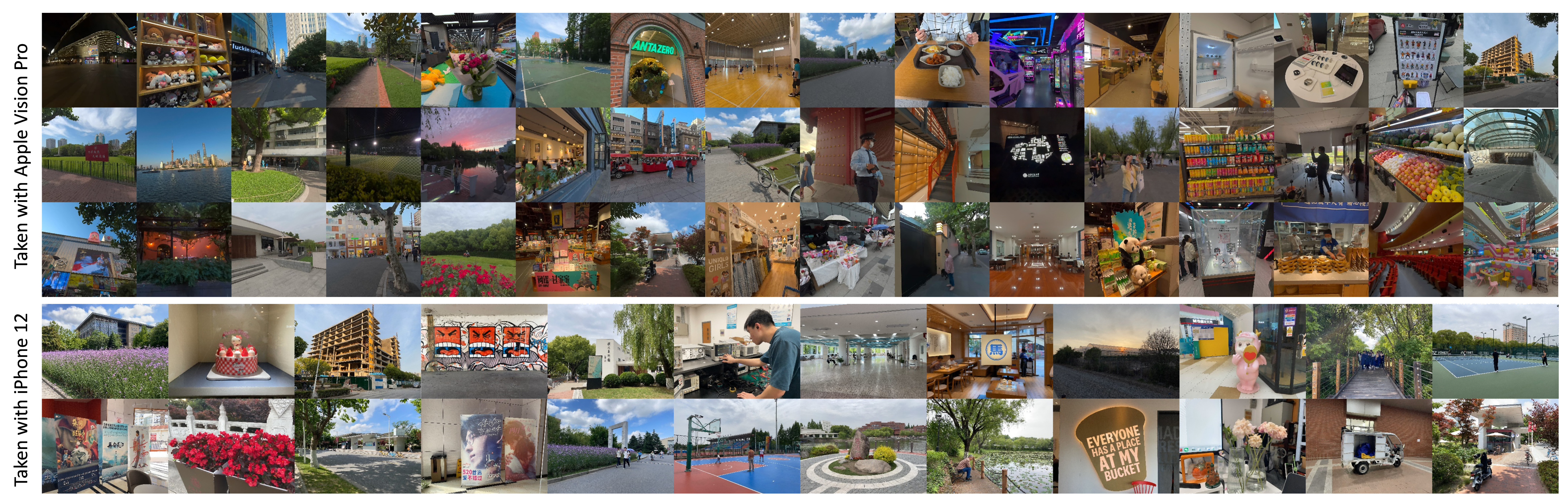}
    \vspace{-10pt}
    \caption{Sample egocentric spatial images from our ESIQAD, where all samples are illustrated in their left view.}
    \label{sample}
    \vspace{-15pt}
\end{figure*}

Egocentric spatial images, taken by a wide variety of individuals from professional photographers to amateurs, exhibit significant differences in visual quality \cite{duan2024quick,min2024perceptual}. 
This quality variation is further compounded by the dynamic and uncontrolled environments when capturing egocentric spatial images  \cite{duan2022confusing}, leading to common distortions such as under/over exposure, low visibility, noise, color shift, \textit{etc.}
Moreover, since the captured egocentric spatial images can be viewed on any device, the corresponding Quality of Experience (QoE) may also be different.
In order to ensure the QoE of end-users, the service providers need to monitor the quality of egocentric spatial images in the entire media stream, including uploading, compressing, post-processing, and transmission. With thousands of egocentric spatial images being captured, uploaded and viewed every day, it is crucial to develop specialized image quality assessment (IQA) metrics for egocentric spatial images to better maintain and optimize the QoE.

Over the past few decades, many studies have focused on the traditional 2D image quality assessment problem, with the establishment of several IQA databases, \textit{e.g.,} TID2013 \cite{ponomarenko2013color}, KonIQ-10K \cite{lin2018koniq}, and KADID-10k \cite{lin2019kadid}, and the development of many advanced IQA models \cite{bmpri, niqe, hosa, ilniqe}.
However, traditional IQA models are inadequate for assessing the quality of egocentric spatial images due to their unique immersive experiences, which have a distinct impact on QoE. Moreover, spatial images consist of two distinct views (left and right), which precludes the direct application of traditional IQA models.
With the diversification of image forms, some studies have focused on 3D IQA (also known as stereoscopic IQA) \cite{fang2019stereoscopic,gorley2008stereoscopic,moorthy2013subjective, zhou2019dual} and VR IQA \cite{duan2023attentive, duan2018perceptual, sun2019mc360iqa,zhu2023perceptual, lim2018}. 
However, existing 3D IQA metrics are mainly designed for traditional stereoscopic images, which rely on dual-eye views to create depth perception. These metrics take 3D effects into account but overlook the immersive experience of egocentric spatial images (which generally requires a larger field of view (FoV) compared to classical 2D/3D small display) and the first-person perspective in dynamic environments. Meanwhile, VR IQA metrics are typically applied to VR images such as omnidirectional images in immersive environments. However, these metrics generally ignore the stereoscopic depth effects and are not designed for the first-person egocentric perspective.
As far as we know, there is a lack of IQA studies for egocentric spatial images, which have significant differences compared with traditional 2D images, 3D images, and VR images in terms of formats, characteristics, and applications. Therefore, there is an urgent need for dedicated research on egocentric spatial image quality assessment from both subjective and objective perspectives.

To better understand human visual preferences for egocentric spatial images and facilitate the development of IQA models tailored to these images, we construct the first \underline{\textbf{E}}gocentric \underline{\textbf{S}}patial \underline{\textbf{I}}mages \underline{\textbf{Q}}uality \underline{\textbf{A}}ssessment \underline{\textbf{D}}atabase, termed ESIQAD. The database contains 500 egocentric spatial images with diverse scenes and the corresponding collected human perceptual quality ratings under three display modes, including 2D display, 3D-window display, and 3D-immersive display. Specifically, 400 egocentric spatial images are captured using the Apple Vision Pro, and 100 egocentric spatial images are taken using the iPhone and synthesized by the iPhone’s ``Spatial Camera" application. Based on the collected images, we conducted subjective experiments to evaluate the QoE of them, in which 22 participants were included to provide their perceptual quality ratings under three display modes. 
Based on our ESIQAD, we introduce a novel mamba2-based multi-stage feature fusion model, which leverages the effectiveness of the state space models (SSM) to predict the perceptual quality of egocentric spatial images across three display modes, termed ESIQAnet. We first input the stereo image pairs into a patch embedding layer. The model then extracts features using visual state space duality (VSSD) blocks in the first three stages and multi-head self attention (MSA) blocks in the final stage. For the extracted multi-scale features, we apply cross attention and transposed attention to fuse and enhance visual information from both views. The outputs are then averaged and concatenated to form a comprehensive feature representation, which is fed into a quality regression network to predict the quality score. 
We validate the performance of the proposed model and 22 state-of-the-art benchmark IQA models on the ESIQAD. Extensive experimental results demonstrate that our model outperforms the benchmark IQA models and has good generalization performance for different display modes.
Our contributions are summarized as follows: \vspace{-1pt}
\begin{itemize}
    \item We establish a large-scale quality assessment database for egocentric images, named ESIQAD, which is the first IQA database for egocentric spatial images to the best of our knowledge. \vspace{-2pt}
    \item We analyze the human preference characteristics for egocentric spatial images under three display modes based on our ESIQAD. \vspace{-2pt}
     \item We conduct a benchmark experiment by evaluating the performance of numerous state-of-the-art IQA models. \vspace{-2pt}
    \item We propose ESIQAnet, which is built based on mamba2 architecture to perform egocentric spatial image quality assessment.
Our ESIQAnet outperforms other benchmark models across all three display modes. 
\end{itemize}

\vspace{-4pt}
\section{Database Construction}
\label{database}
\vspace{-2pt}
To facilitate the egocentric spatial IQA research, we construct a large-scale database, termed ESIQAD, and conduct the corresponding subjective experiment to derive the subjective quality scores in terms of three display modes.
\vspace{-4pt}
\subsection{Content Collection}
\vspace{-2pt}
Firstly, we utilized the Apple Vision Pro and the iPhone 12 as the shooting devices to collect egocentric spatial images from a variety of scenes, including both indoor and outdoor environments, different weather conditions, and various times of day. These images feature various characteristics including brightness, contrast, colorfulness, spatial details, \textit{etc.}, and have different degradations including noise, blur, \textit{etc.} The raw images were taken by ten photographers, representing a wide range of user perspectives and preferences. 
Specifically, we collected 400 egocentric spatial images captured with the Apple Vision Pro, and 100 egocentric spatial images synthesized from the images taken by the iPhone, referred to as \textit{captured images} and \textit{synthesized images}, respectively. 
For the 100 synthesized images, we used the wide and ultra-wide cameras of iPhone 12 to simultaneously capture two images, serving as the left and right views for synthesizing a spatial image. When the iPhone is held horizontally, the two cameras can mimic human eye parallax.
It should be noted that the 100 synthesized images have their corresponding images with the same scenes in the 400 images captured by Apple Vision Pro to facilitate the study of human visual preferences.
The captured images have a fixed resolution of 2560 × 2560 per view, while the synthesized images have a fixed resolution of 4032 × 3024 per view. \cref{sample} demonstrates sample images from our ESIQAD, where all samples are illustrated in the main view (left view). 
\vspace{-2pt}
\subsection{Subjective Experiment}
\vspace{-2pt}

\begin{figure}[t]
\vspace{-10pt}
  \centering
  \subfloat[2D]{
        \includegraphics[width=0.345\linewidth]{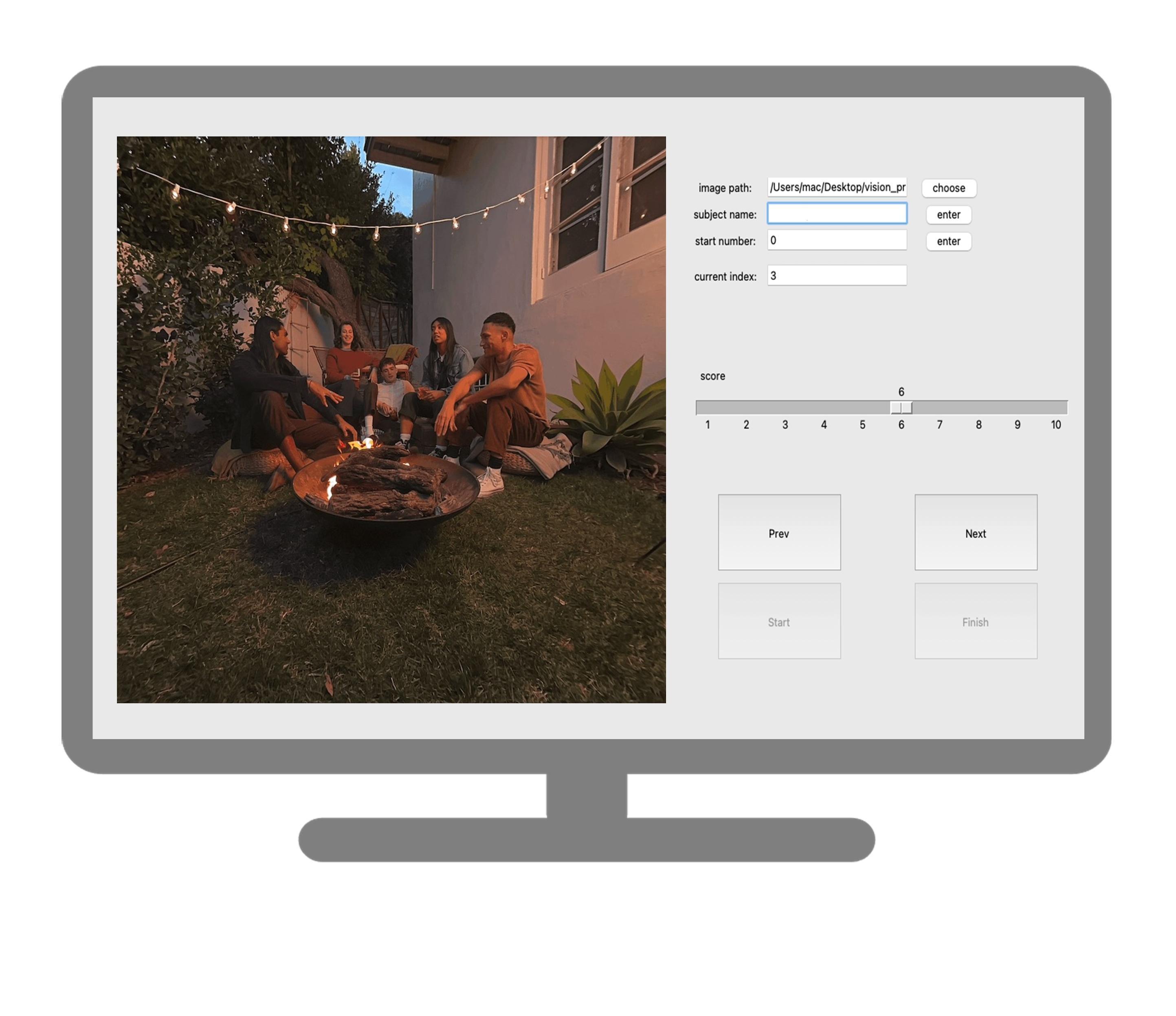}}
    \subfloat[3D-window]{
        \includegraphics[width=0.305\linewidth]{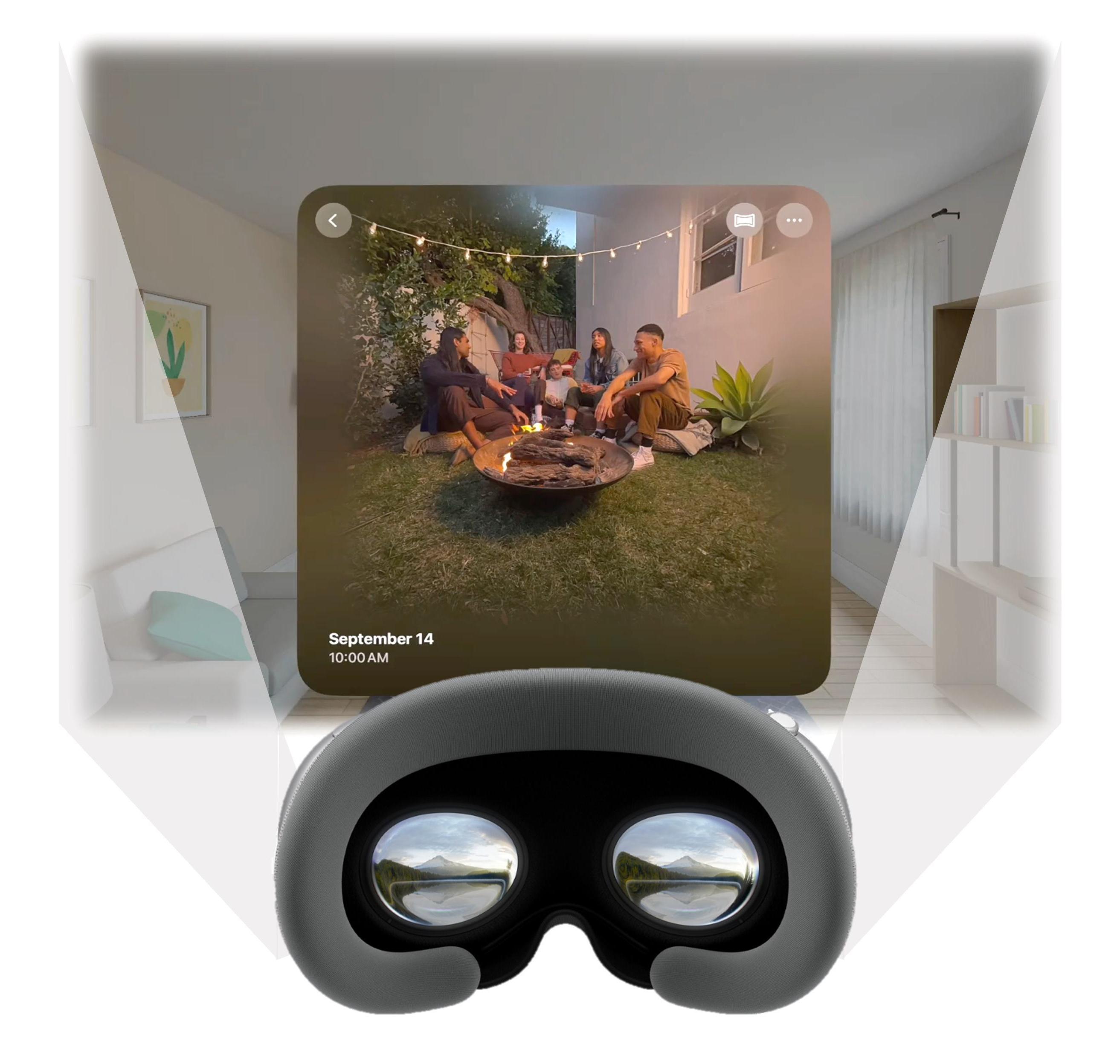}}
    \subfloat[3D-immersive]{
        \includegraphics[width=0.345\linewidth]{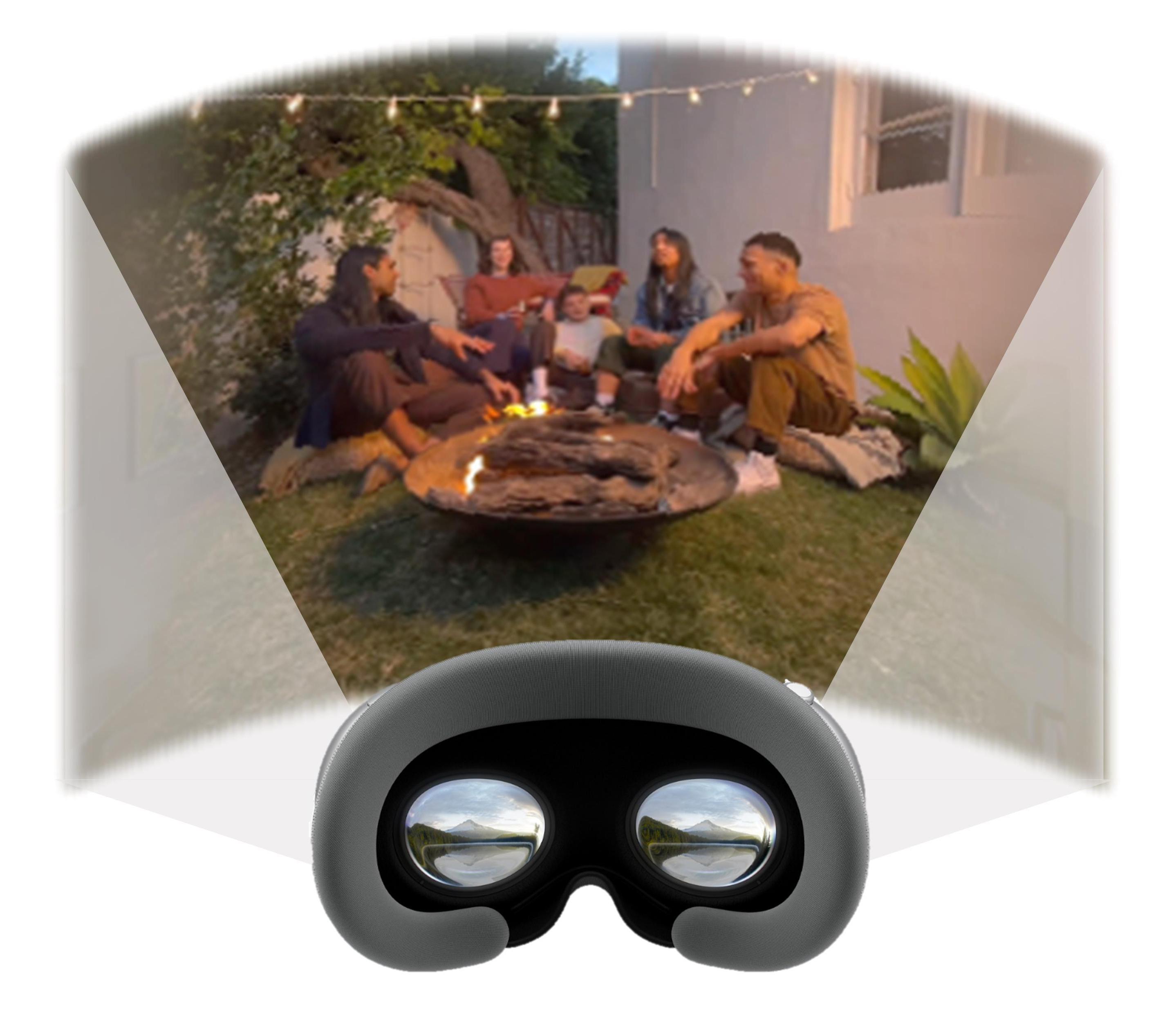}}
  \vspace{-10pt}
  \caption{Three display modes in our subjective experiments.}
    \vspace{-18pt}
  \label{modes}
\end{figure}
\subsubsection{Experiment setup}
\vspace{-2pt}
Based on the collected images, we then conducted a subjective experiment to gather quality scores of human perception on egocentric spatial images across different display modes, including 2D, 3D-window, and 3D-immersive, as shown in \cref{modes}. 
For the 2D display mode, egocentric spatial images were displayed on an iMac monitor with a resolution of 4K, driven by a Mac Mini M2 PC. We developed a quality scoring interface using Python Tkinter, which was used to present egocentric spatial images (only the left view) and collect subjective quality ratings. The interface allows users to freely browse the previous and next images, and record the quality scores through a scroll bar.
For the 3D-window and 3D-immersive display modes, we utilized the Apple Vision Pro device as the HMD, which features a resolution of 3660 × 3200 per eye, a 120$\degree$ horizontal FoV, and a 100Hz refresh rate. For the 3D-window mode, egocentric spatial images appeared as small windows within the HMD, allowing users to maintain awareness of the surroundings. For the 3D-immersive mode, egocentric spatial images were shown in a wider immersive FoV, which provides an immersive egocentric experience. In both 3D display modes, all images were displayed through the built-in application ``Photos'' and displayed at their native resolutions to prevent the scaling distortion. Across all three display modes, egocentric spatial images were presented in a random order. 

\vspace{-4pt}
\subsubsection{Experiment methodology}
\vspace{-2pt}

We adopted the single stimulus absolute category rating (SSACR) method to collect subjective quality ratings for egocentric spatial images under three display modes, with the rating scale ranging from 1 to 10. 
Under the 2D display, the viewers were seated at a distance of about 2 feet from the monitor in a laboratory environment with normal indoor illumination and utilized the scoring interface to assign quality scores. 
Under the two 3D displays, participants were instructed to wear the Apple Vision Pro and select either the 3D-window or 3D-immersive display mode via the ``Photos" app. A keyboard was placed
in front of the participants to enter the scores. Participants manually entered their scores after viewing each image and proceeded to the next image using gestures.



According to the recommendation from ITU-R BT.500-13 \cite{itu2012bt500} and BT.2021 \cite{itu2012bt2021}, we invited a total of 22 graduate students to participate in the subjective experiment. Before the experiment, all participants completed a survey to provide the participant demographics.
Among all participants (11 males and 11 females, age range 22–30, average age 25), 14 indicated that they spend no time in HMDs per week, while 6 indicated that they spend between 0.5-3 hours, and 2 indicated they spend > 3 hours.
All participants were required to complete a standard procedure in visual testing, \textit{i.e.}, the Snellen visual acuity test \cite{SnellenChart2020}, where 8 of them were confirmed to have normal vision and 14 had corrected-to-normal vision (9 wore contact lenses, 2 wore glasses, 2 wore neither, and 1 had undergone refractive surgery). 
For all three display modes, the participants were instructed to assess the egocentric spatial images based on both distortion and aesthetic quality, providing ratings with an overall quality score. The whole experiment was split into 2 sessions, each of which included a subjective quality evaluation process for 750 samples (250 spatial images $\times$ 3 display modes). All participants completed both sessions, producing a total of 33,000 quality ratings (22 participants $\times$ 1,500 ratings). After the scoring sessions, each participant was asked to answer a post-study questionnaire regarding their experience. Details of the questionnaire are given in \cref{2.3.2}.

\begin{figure}
\vspace{-10pt}
    \centering
    \hspace{-4pt}\subfloat[MOS discriminability]{
        \includegraphics[width=0.47\linewidth]{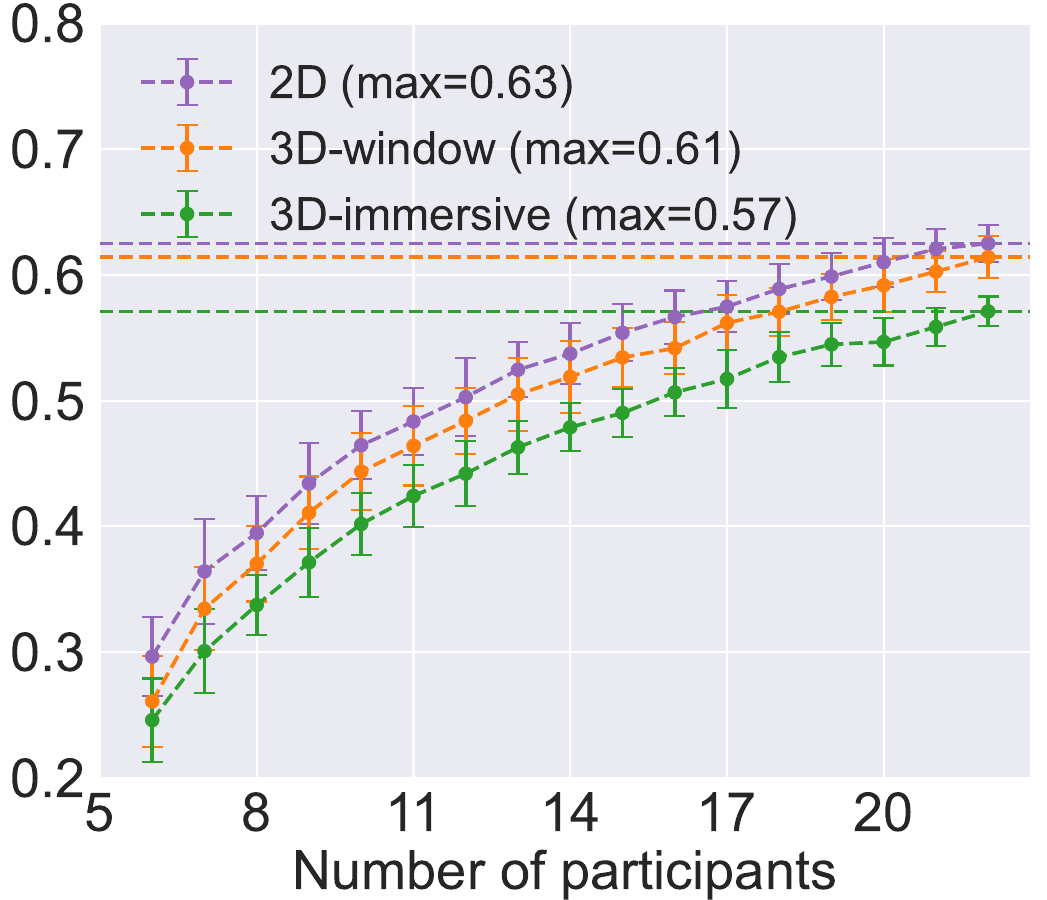}}
    \subfloat[Mean CI]{
        \includegraphics[width=0.479\linewidth]{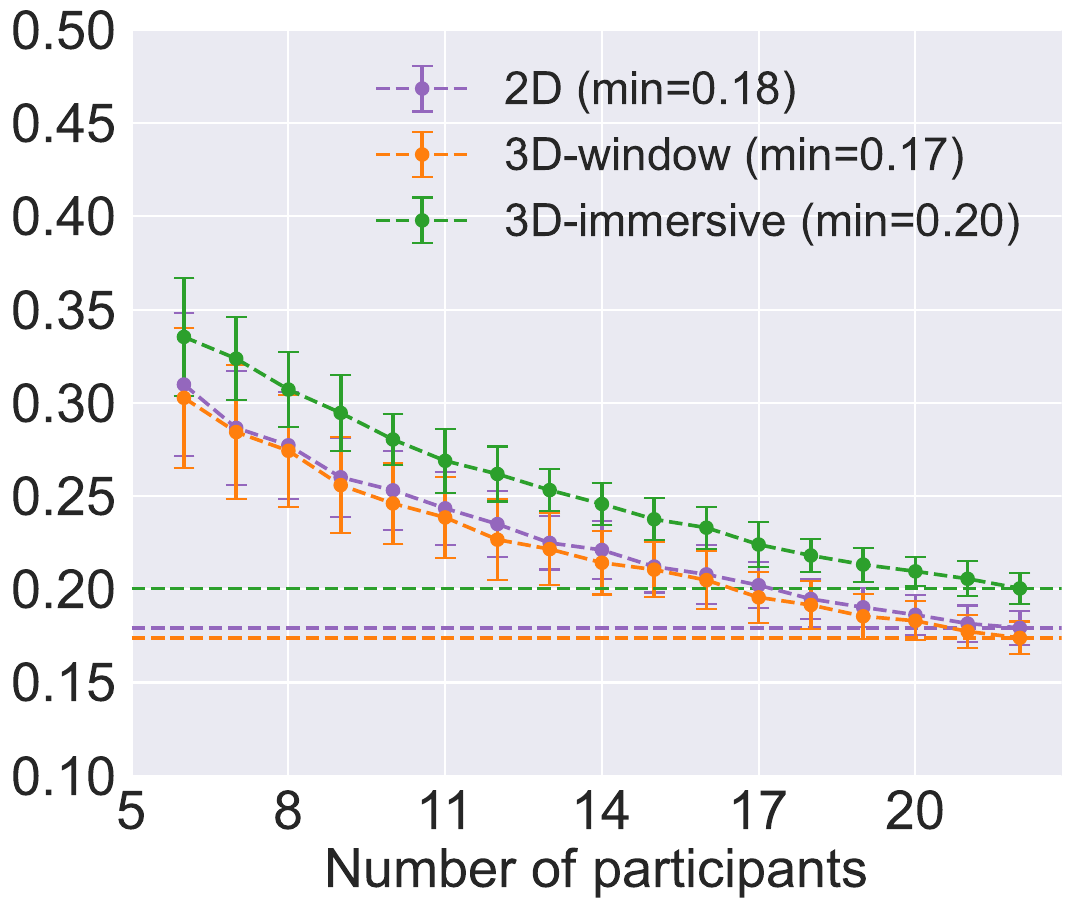}}
        \vspace{-10pt}
    \caption{MOS discriminability and mean CI evolution with participants’ number in ESIQAD.}
        \vspace{-8pt}
    \label{discriminability}
\end{figure}

\begin{figure}[!ht]
    \centering
\subfloat[MOS: All images]{ \vspace{-4pt}
\includegraphics[width=0.75\linewidth]{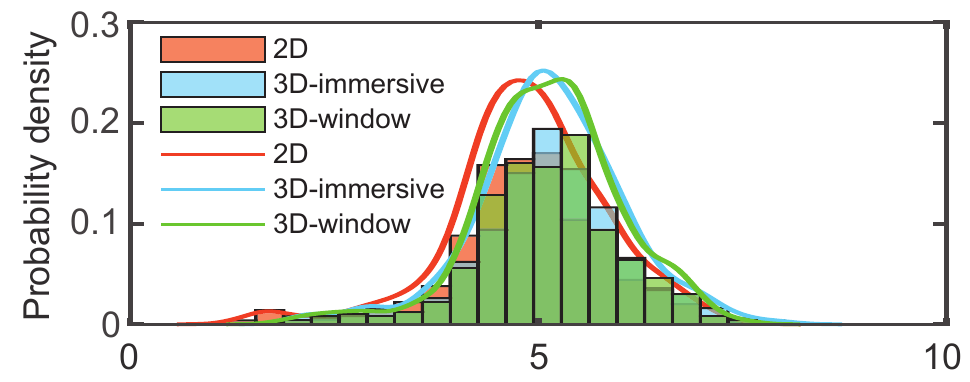}}\\
\vspace{-2pt}
\subfloat[MOS: Captured images]{ \vspace{-5pt}
\includegraphics[width=0.75\linewidth]{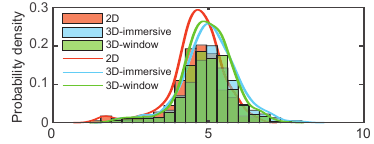}}\\
\vspace{-2pt}
\subfloat[MOS: Synthesized images]{ \vspace{-6pt}
\includegraphics[width=0.75\linewidth]{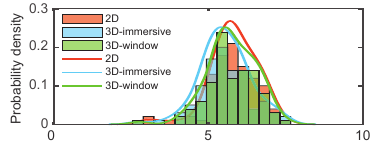}}\\
        \vspace{-8pt}
    \caption{Distribution of MOSs of egocentric spatial images across three display modes.}
        \vspace{-15pt}
    \label{MOS}
\end{figure}

\subsection{Subjective Data Processing and Analysis}
\subsubsection{MOS Calculation and Analysis} \label{Analysis}
We performed the same processing methods for the collected scores under three display modes respectively. Firstly, we followed the subjective data processing guidelines recommended by ITU \cite{ITU-R_BT_500-13} for outlier detection and subject rejection. 
For 2D and 3D-immersive display modes, 1 of the 22 participants was deemed as an outlier and excluded, respectively. For the 3D-window display mode, none of the 22 participants were excluded.
Then, we convert the raw scores from the subjects into normalized Z-scores, which range between 0 and 100. Subsequently, we calculate the average of these Z-scores to derive the mean opinion scores (MOSs), which are formulated as follows: 
\begin{equation}
\setlength{\abovedisplayskip}{3pt}
\setlength{\belowdisplayskip}{0pt}
    z_{i j}  =\frac{r_{i j}-\mu_i}{\sigma_i}, \quad z_{i j}^{\prime}=\frac{100\left(z_{i j}+3\right)}{6},
\end{equation}
\begin{equation}
\setlength{\abovedisplayskip}{-3pt}
\setlength{\belowdisplayskip}{0pt}
    \text{MOS}_j  =\frac{1}{N} \sum_{i=1}^N z_{i j}^{\prime},
\end{equation}
where $r_{ij}$ is the original score of the $i$-th subject on the $j$-th image, $\mu_i$ and $\sigma_i$ represent the mean rating and the standard deviation given by subject $i$, respectively, and $N$ is the total number of subjects.

To investigate the reliability of MOSs, we calculated the \textit{discriminability} and \textit{mean Confidence Interval (CI)} metrics of the ESIQAD with increasing assessor numbers \cite{pastor2023towards,10448123}. For the discriminability metric, we applied the two-sample Wilcoxon test on all possible pairs of MOSs in the ESIQAD to test the ratio of significantly different ones according to \cite{10448123}. 
The mean CI was calculated by averaging the standard deviations of scores for each sample (scaled by a Z-score for a 95\% confidence level) across all samples, which measures the average uncertainty around the MOSs within the database. \cref{discriminability} illustrates the trend of the discriminability and mean CI metrics as the number of participants changes, showing that 22 participants can yield relatively reliable MOSs.




\begin{figure}[t]
    \begin{minipage}[t]{0.98\linewidth}
    \vspace{-10pt}
        \centering
        \includegraphics[width=\linewidth]{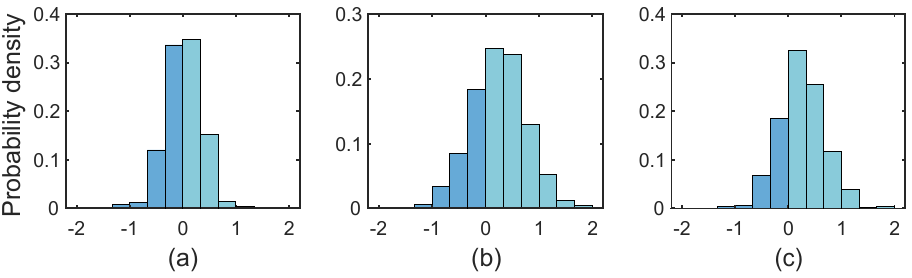}
        \vspace{-16pt}
        \caption{Distribution of MOS difference of spatial images across three modes. (a) $\text{MOS}_\text{3D-window}-\text{MOS}_\text{3D-immersive}$. (b) $\text{MOS}_\text{3D-immersive}-\text{MOS}_\text{2D}$. (c) $\text{MOS}_\text{3D-window}-\text{MOS}_\text{2D}$.}
        \label{MOS_difference}
    \end{minipage} \vspace{4pt} \\
    \begin{minipage}[t]{0.98\linewidth}
        \centering
        \includegraphics[width=\linewidth]{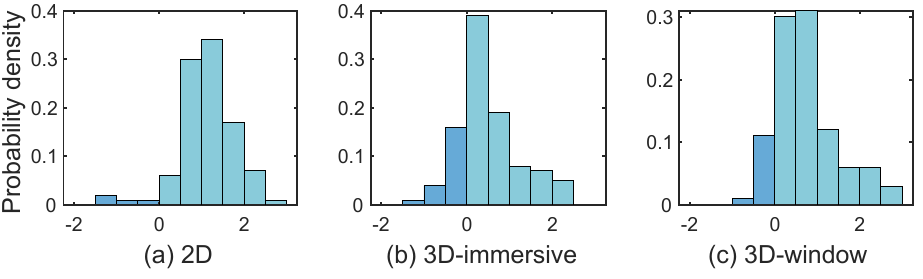}
        \vspace{-16pt}
        \caption{Distributions of the MOS difference ($\text{MOS}_{\text{synthesized}}-\text{MOS}_{\text{captured}}$) between 100 matched pairs of captured and synthesized images under three display modes.}
        \label{MOS_difference2}
        \vspace{-16pt}
    \end{minipage}
\end{figure}

\cref{MOS}(a) presents the histograms of the MOS distribution over the three viewing modes, indicating that the perceptual quality scores are widely distributed in the $\left[1,10\right]$ interval in each mode. 
In addition, \cref{MOS}(b) and \cref{MOS}(c) demonstrate the MOS distributions of 400 captured images and 100 synthesized images, respectively. It can be observed that the MOS distribution for captured images centers around 5 in all three modes, while it centers above 5 for synthesized images. 
This is mainly because the image captured by the iPhone has a higher fixed resolution and their color is generally enhanced to be more comfortable for human vision perception.
Moreover, for captured images, the perceptual quality is higher in 3D-immersive and 3D-window viewing modes compared to the 2D viewing mode. However, for synthesized images, the perceptual quality is better in 2D viewing mode than in 3D viewing modes.

The results shown in \cref{MOS}(a) indicate that the MOS fitting curves for the two 3D viewing modes are located at the right of the 2D viewing mode, manifesting higher perceived image quality in the 3D environments. 
To facilitate a more direct comparison of the perceptual quality of the same image across different viewing modes, we calculate the MOS difference for each image between the three modes. The differences are illustrated in the frequency distribution histograms as shown in  \cref{MOS_difference}.
The results indicate that the perceptual quality of most images in 3D-window and 3D-immersive scenes is comparable but generally superior to that observed on 2D displays.
This may be attributed to the immersive and stereoscopic nature of 3D viewing modes, which offer better senses with depth and realism. In 3D environments, viewers experience enhanced spatial details and a more real scene representation, leading to an overall improvement in perceived image quality compared to 2D displays.

To compare the perceptual quality of captured and synthesized images with identical scenes, we compute the MOS difference between 100 matched pairs of captured and synthesized images across three display modes by subtracting the MOS of synthesized images and their corresponding captured images. The distribution histogram of the differences is depicted in \cref{MOS_difference2}. 
The results indicate that the perceptual quality of most synthesized images exceeds that of their corresponding captured images. 
This improvement mainly results from the higher resolution and more accurate color reproduction in synthesized images taken by phones.

\begin{figure}[t]
\vspace{-10pt}
    \centering
    \includegraphics[width=0.98\linewidth]{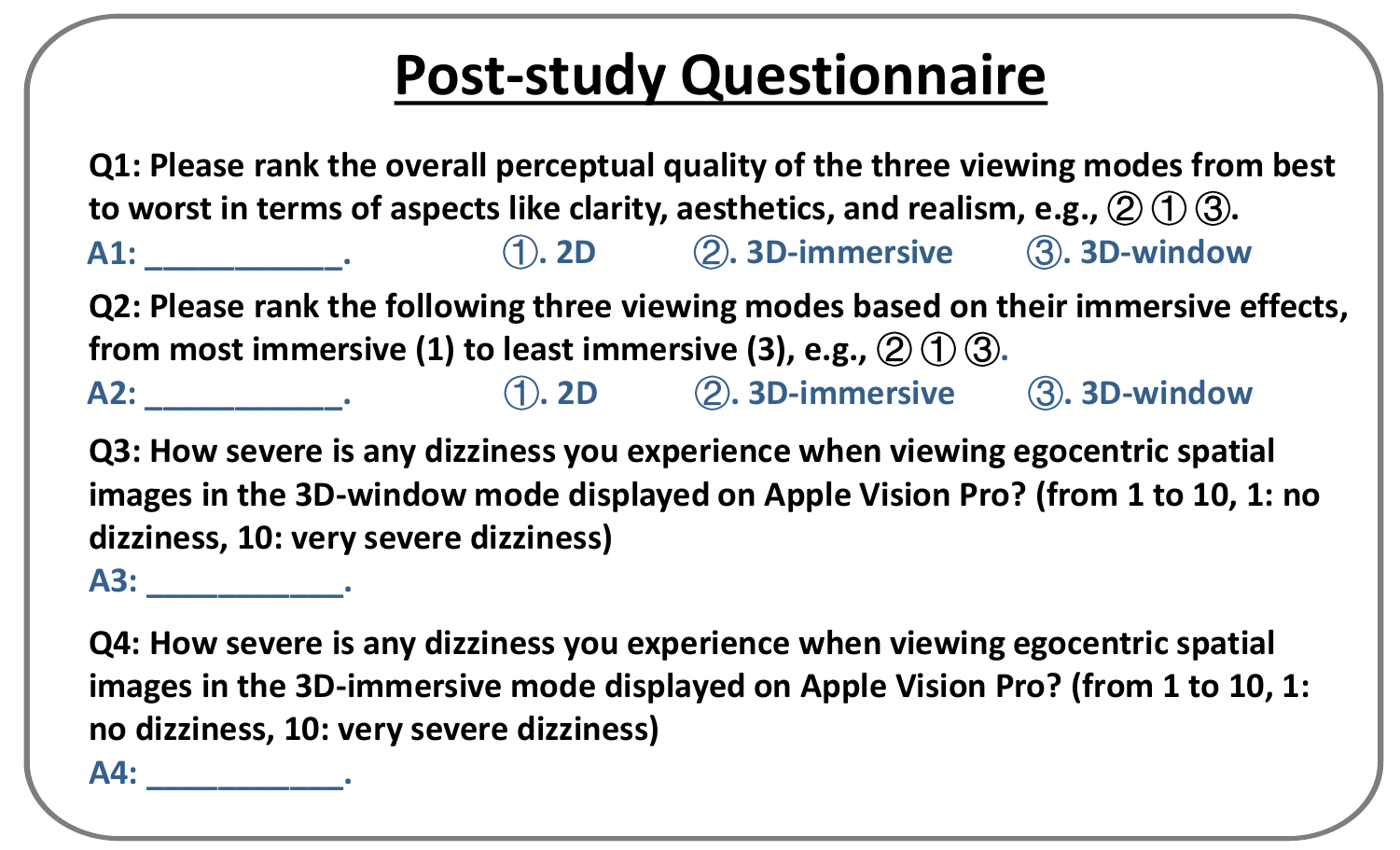}
            \vspace{-6pt}
    \caption{Questions included in the post-test questionnaire.}
    \label{questionnaire} \vspace{-9pt}
\end{figure}

\begin{figure}
    \centering
    \includegraphics[width=0.55\linewidth]{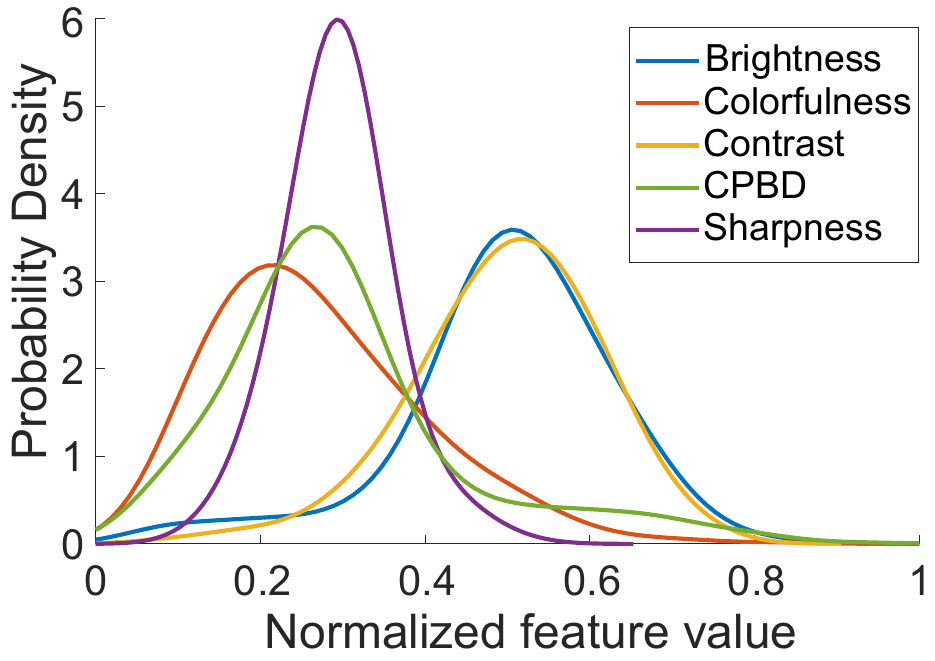}
    \vspace{-6pt}
    \caption{Distributions of five low-level vision features in our ESIQAD.}
    \vspace{-18pt}
    \label{fig:kernel}
\end{figure}

\begin{figure*}[t]
\vspace{-15pt}
    \centering
\includegraphics[width=0.995\linewidth]{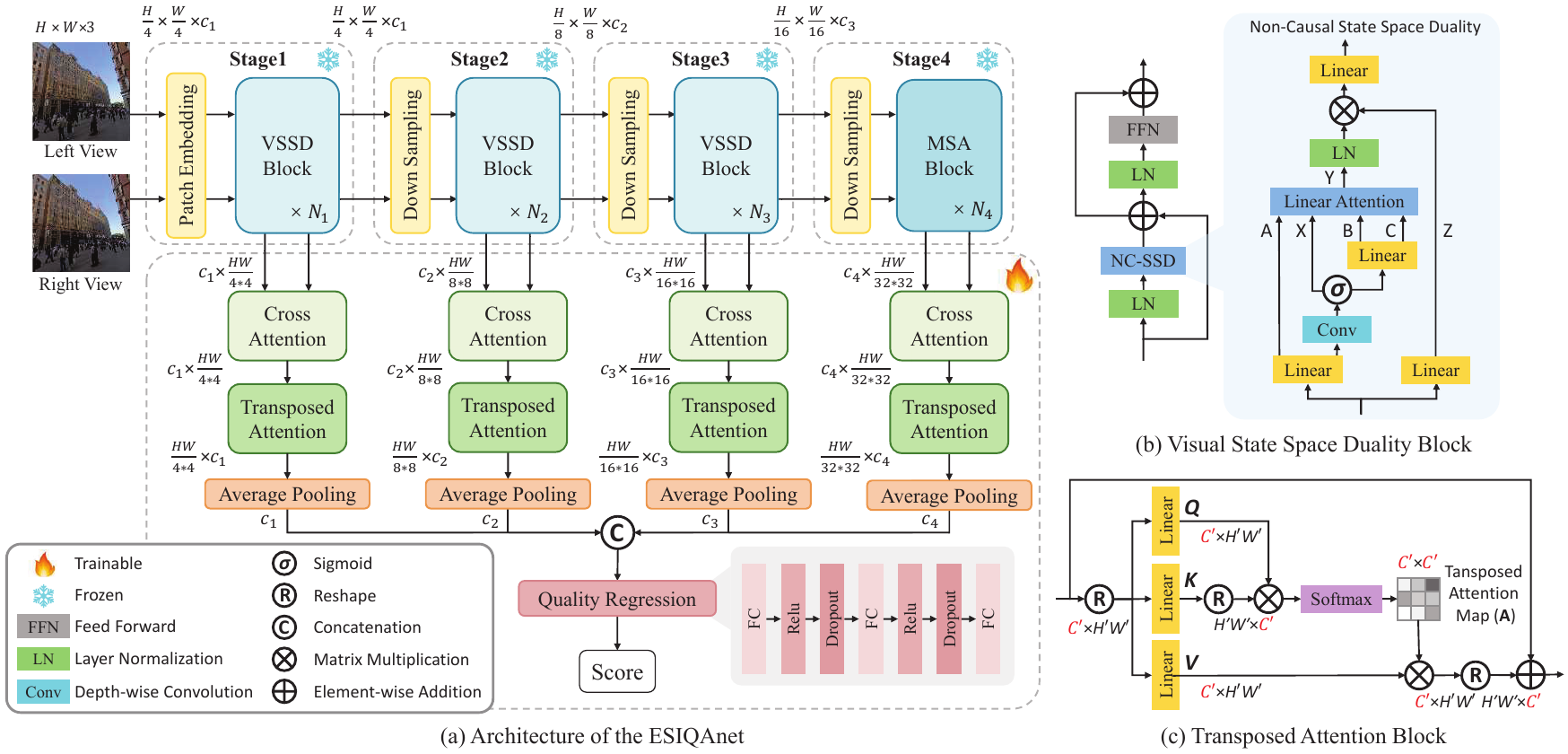}
\vspace{-5pt}
    \caption{Illustration of (a) the overall architecture of the ESIQAnet, (b) the structure of visual state space duality (VSSD) block, and (c) the structure of transposed attention block.}
    \label{framework}
    \vspace{-14pt}
\end{figure*}  

\vspace{-2pt}
\subsubsection{Questionnaire Analysis}\label{2.3.2}
\vspace{-2pt}
Post-test questionnaires can help better understand the QoE of egocentric spatial images and design future studies. All questions are shown in  \cref{questionnaire}. For Q1 and Q2, we derive ranking scores for the three display modes based on the sorting results, which reflect the overall ranking of the options.
The ranking score for a display mode is calculated by 
\begin{equation}
\vspace{-4pt}
S = \frac{\sum (f \times w)}{n},
\vspace{-4pt}
\end{equation}
where $S$ indicates the ranking score for this display mode, $f$ represents the frequency of each rank, $w$ denotes the weight assigned to each rank (\textit{e.g.,} the top rank is weighted 3, and the lowest rank is weighted 1), and $n$ is the total number of participants. 
For Q3 and Q4, we compute the average scores from questionnaire responses to evaluate user dizziness.
The statistical results of the replies and some general observations are as follows: (1) Q1: 3D-immersive mode, 3D-window mode, and 2D mode obtain ranking scores of 2.27, 2.18, and 1.55, respectively. The results suggest that most subjects prefer the experience of 3D-immersive and 3D-window modes viewing in Apple Vision Pro compared to the experience of the 2D mode viewing on the screen display, which is consistent with our previous analysis in \cref{Analysis}. (2)  Q2: We derive the ranking scores for each display mode: 3.00, 1.73, and 1.14 for 3D-immersive, 3D-window, and 2D modes, respectively. The results indicate participants generally consider that the 3D-immersive scene offers the most immersive experience. (3) Q3 \& Q4:  The scores for dizziness in the 3D-immersive and 3D-window display modes are 4.27 and 2.27, respectively. This result suggests that while the 3D-immersive mode offers a superior immersive experience, it also induces more significant dizziness compared to the 3D-window mode.

\vspace{-2pt}
\subsubsection{Database Attributes Analysis}
\vspace{-2pt}
Following the methodologies outlined in \cite{hansen1998effective} and \cite{winkler2012analysis}, we conduct statistical analysis for our ESIQAD in terms of five low-level vision feature dimensions, including sharpness, brightness, colorfulness, contrast, and cumulative probability of blur detection metric (CPBD). Considering that the ranges of these features are different, we normalized their values to $\left[0, 1\right]$ for better illustration.
\cref{fig:kernel} shows the kernel distribution of each feature for our ESIQAD. It can be observed that the spatial images in our database have a wide distribution in brightness, colorfulness, contrast, and CPBD features, showing their extensive diversity. However, our database shows a narrow distribution range for the sharpness feature due to the fixed resolutions. 

\vspace{-2pt}
\section{Proposed Method}
\subsection{Overall Pipeline}
\vspace{-2pt}

This section introduces our proposed model, which is designed to predict the quality scores of egocentric spatial images under three display modes. Consider a spatial image $\textbf{\textit{I}}$, consisting of a left view $\textbf{\textit{I}}_{\text{\textit{l}}} \in \mathbb{R}^{H \times W \times 3}$ and a right view $\textbf{\textit{I}}_{\text{\textit{r}}} \in \mathbb{R}^{H \times W \times 3}$, where $H$ and $W$ represent the image height and width, respectively. The stereo image pairs are processed by patch embedding and then forwarded to multi-stage hierarchical VSSD blocks and a final stage of multi-head self attention (MSA) blocks for spatial feature extraction. 
We denote the spatial features from the $i{\text{-th}}$ stage as $\textbf{\textit{V}}^{\textit{l}}_i$ and $\textbf{\textit{V}}^{\textit{r}}_i \in \mathbb{R}^{b \times c_i \times H_i W_i }$, 
where $i \in\{1, 2, \dots, n\}, b$ denotes the batch size, $c_i$, $H_i$, and $W_i$ denote the channel size, width, and height of the $i{\text{-th}}$ feature, respectively. 
At each stage, the extracted features from the left and right views, $\textbf{\textit{V}}^{\textit{l}}_i$ and $\textbf{\textit{V}}^{\textit{r}}_i \in \mathbb{R}^{b \times c_i \times H_i W_i}$, are fed into a cross attention block to generate interactive content-aware feature maps $\textbf{\textit{F}}_i \in \mathbb{R}^{b \times c_i \times H_i W_i}$.
We employ a transposed attention block to enhance channel interactions in $\textbf{\textit{F}}_i$, producing the deep visual feature $\tilde{\textbf{\textit{F}}}_i$. Then we apply average pooling to $\tilde{\textbf{\textit{F}}}_i$ to aggregate spatial information, generating a channel-level global representation. Finally, we concatenate the deep visual features from all stages and use a quality regression module to predict the final perceptual quality score. The architecture for the ESIQAnet is depicted in \cref{framework}(a).
\vspace{-5pt}
\subsection{Visual State Space Duality Block}
\vspace{-2pt}
The visual state space duality (VSSD) block develops from the state space model (SSM), known for its efficiency in modeling sequential data \cite{gu2023mamba}, which can be formulated as:
\begin{equation}
\setlength{\abovedisplayskip}{3pt}
\setlength{\belowdisplayskip}{2pt}
    h(t)=Ah(t-1)+Bx(t), \quad y(t)=Ch(t),
\end{equation}
where $A$, $B$, and $C$ are learned parameters defining the system's state transitions. 
Mamba2 \cite{dao2024transformers} leverages both the linear recurrence of SSM and the quadratic dual form, introducing state space duality (SSD), which is based on block decompositions of semiseparable matrices,
To enhance the SSD block in Mamba2 for vision applications, the visual state space duality (VSSD) block utilizes a bidirectional scanning strategy to eliminate causality, replacing the SSD with an NC-SSD \cite{shi2024vssd}.
Compared with SSD, VSSD has implemented many improvements. Specifically, the causal convolution 1D in the NC-SSD block is replaced with a depth-wise convolution (DWConv) with a kernel size of 3. Aligning with classical vision transformers \cite{dosovitskiy2020vit, liu2021swin}, the VSSD block also integrates a feed-forward network (FFN) after the NC-SSD to improve channel communication. Additionally, a local perception unit (LPU) \cite{guo2022cmt} is added before the NC-SSD block and FFN to improve local feature detection.



\begin{figure}[t]
\vspace{-3pt}
    \centering
\includegraphics[width=0.8\linewidth]{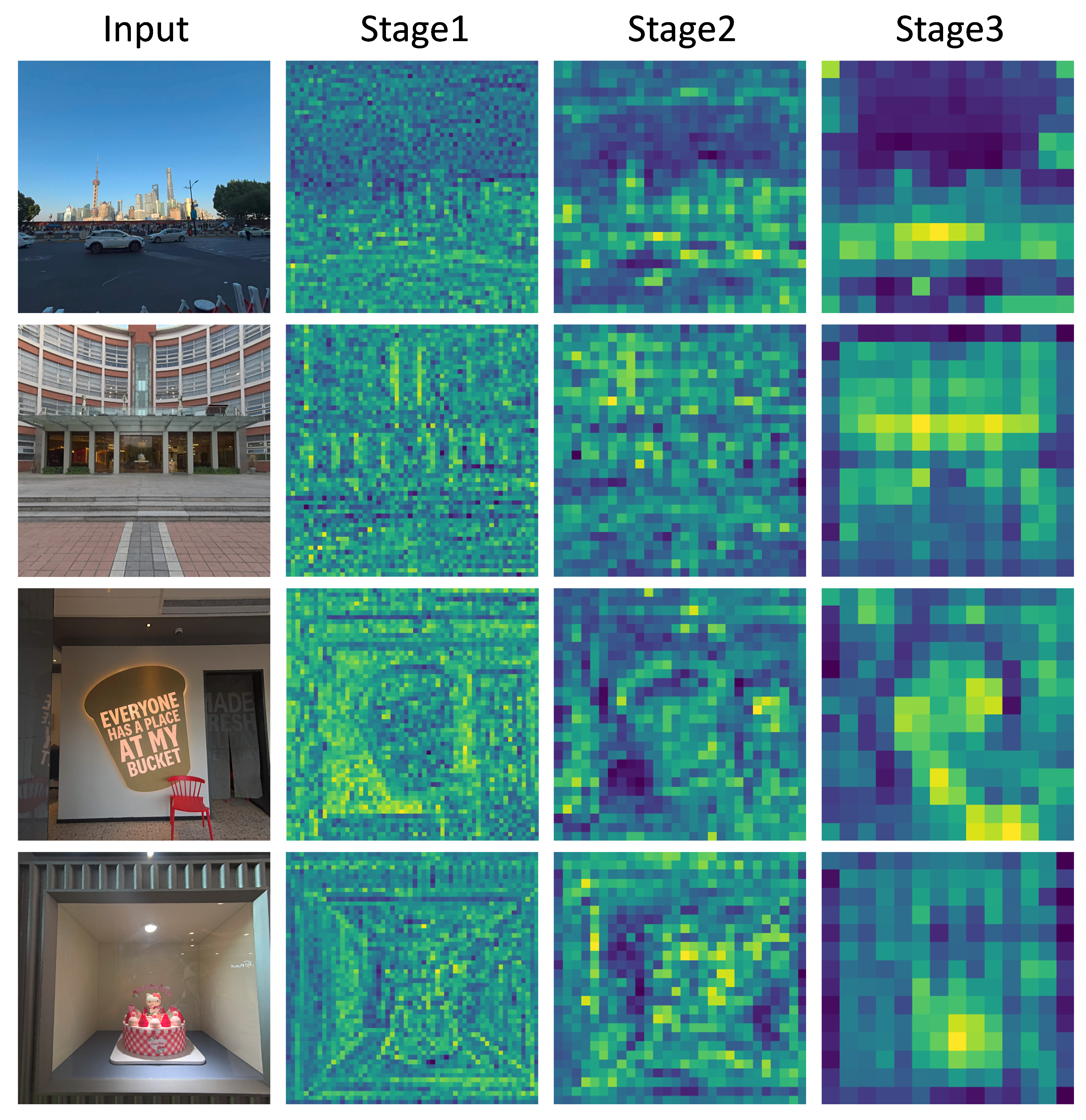}
\vspace{-7pt}
    \caption{Visualization of input images and their corresponding heatmaps obtained from different stages of VSSD blocks, which are derived by averaging the vector $\textbf{\textit{V}}^{\textit{l}}_i$ across various channels.}
    \vspace{-15pt}
    \label{heatmap}
\end{figure}

In our model, we employ multi-stage VSSD blocks to hierarchically extract spatial features from binocular views of egocentric spatial images. The architecture of the VSSD block is depicted in \cref{framework}(b).
In the VSSD block, the input $\textbf{\textit{X}}$ is processed through linear layers and convolutions to obtain the parameters $[A, B, C, \Delta, Z]$, which define the state transition and output behavior of the system. 
The VSSD process is simply formulated as follows:
\begin{equation}
\vspace{-2pt}
   \operatorname{NC-SSD}(\textbf{\textit{X}}) =  \operatorname{LinearAttn}(\Delta, A, B, C) \cdot Z, 
\vspace{-2pt}
\end{equation}
\begin{equation}
\vspace{-2pt}
\textbf{\textit{X}}' = \textbf{\textit{X}} + \operatorname{NC-SSD}(\textbf{\textit{X}}),
\vspace{-2pt}
\end{equation}
\begin{equation}
\vspace{-2pt}
\tilde{\textbf{\textit{X}}}= \textbf{\textit{X}}' + \operatorname{FFN}(\operatorname{LN}(\textbf{\textit{X}}')),
\vspace{-2pt}
\end{equation}
where $\textbf{\textit{X}}$ and $\tilde{\textbf{\textit{X}}}$ represent the input and output feature respectively, and $\operatorname{LN}$ and $\operatorname{FFN}$ represent the feed-forward operation and layer normalization operation, respectively.
In our model, the spatial features output $\textbf{\textit{V}}^{\textit{l}}_i$ and $\textbf{\textit{V}}^{\textit{r}}_i$ by $i{\text{-th}}$ stage are further down-sampled and fed to the next stage which consists of $N_{i+1}$ VSSD blocks, resulting to $H_{i+1} = H_i/2$ and $W_{i+1} = W_i/2$. 

To illustrate the effectiveness of the multi-stage VSSD blocks intuitively, we visualize the feature maps obtained from different stages of VSSD blocks as heatmaps in \cref{heatmap} (using the $\textbf{\textit{V}}^{\textit{l}}_i$ as an example). The results indicate that the multi-stage VSSD blocks progressively extract semantic features of the input image, allowing the model to focus on elements that are most critical for quality assessment.

\vspace{-3pt}
\subsection{Attention mechanisms}
\subsubsection{Cross Attention Block}
\vspace{-2pt}

Cross attention mechanisms are widely used in various vision tasks due to their ability to effectively capture relationships between different inputs \cite{chen2021crossvit, duan2022confusing, lin2022cat}. In our model, we leverage cross attention to mimic the human brain's cognitive preference process of integrating visual information from both eyes to perceive egocentric spatial images. This mechanism simulates how the brain forms preferences and creates an immersive experience of these images under 3D modes. Specifically, we use the cross attention module to fuse features extracted from the left and right views, capturing the disparity information between them to derive interactive features. 

The cross attention block works by computing interactions between a query from one view and keys from the other view. Specifically, for the left and right view features $\textbf{\textit{V}}^{l}_i$ and $\textbf{\textit{V}}^{r}_i$, the cross attention module generates \textit{query} ($\textbf{\textit{Q}}$), \textit{key} ($\textbf{\textit{K}}$), and \textit{value} ($\textbf{\textit{V}}$) projections through linear transformations of the input features. The query from one view interacts with the keys of the other view via a dot-product operation to produce an attention map that reflects the correlations between the views, where the attention map $\textbf{\textit{A}}$ can be formulated as:
\begin{equation}
\setlength{\abovedisplayskip}{3pt}
\setlength{\belowdisplayskip}{2pt}
\textbf{A}=\operatorname{Softmax}\left(\frac{\textbf{\textit{Q}}^l\left(\textbf{\textit{K}}^r\right)^T}{\sqrt{d_k}}\right),
\end{equation}
where $\textbf{\textit{Q}}^{l}$ and $\textbf{\textit{K}}^{r}$ are the query and key projections from the left and right views, respectively, and $d_k$ is the scaling factor. This attention map is then used to weight the value projection $\textbf{\textit{V}}^{r}_i$, aggregating information from the right view into the left view's context. 

\vspace{-2pt}
\subsubsection{Transposed Attention Block}

Traditional self attention employs key-query dot-product interactions to establish global connections among spatial patches but overlooks channel-wise information \cite{vaswani2017attention}. To address this issue, the transposed attention block applies self attention across channels rather than the spatial dimension  \cite{maniqa, zamir2022restormer}. This method calculates cross-covariance across channels to produce an attention map that implicitly encodes global context. 


In our model, we utilize the transposed attention block to prioritize the channels of the integrated spatial features $\textbf{\textit{F}}_i$ and enhance downstream local interactions by leveraging the globally contextual information encoded by the attention map. 
Specifically, the transposed attention generates \textit{query} ($\textbf{\textit{Q}}$), \textit{key} ($\textbf{\textit{K}}$), and \textit{value} ($\textbf{\textit{V}}$) projections via three independent linear projections of $\textbf{\textit{F}}_i$. Then we reshape the query and key projections for their dot-product interaction, producing a transposed attention map $\textbf{A} \in \mathbb{R}^{c'\times c'}$. The architecture of the transposed attention block is depicted in \cref{framework}(c), which can be formulated as:

\begin{equation}
\setlength{\abovedisplayskip}{1pt}
\setlength{\belowdisplayskip}{0pt}
\textbf{A} = \operatorname{Softmax}\left(\frac{\textbf{\textit{Q}}\textbf{\textit{K}}}{\sqrt{d_k}}\right),
\end{equation}
where $d_k$ is the dimension of each head in the attention mechanism.
Then we obtain the transposed attention output feature $\tilde{\textbf{\textit{F}}}_i$ by applying the residual connection with $\textbf{\textit{F}}_i$:
\begin{equation}
\setlength{\abovedisplayskip}{3pt}
\setlength{\belowdisplayskip}{2pt}
\tilde{\textbf{\textit{F}}}_i=\textbf{\textit{W}}_p \cdot  ( \textbf{\textit{V}} \cdot \textbf{A} ) +\textbf{\textit{F}}_i,
\end{equation}
where $\textbf{\textit{W}}_p$ represents a trainable linear projection matrix. 

\vspace{-2pt}
\subsection{Quality Regression}
\vspace{-2pt}

The quality regression module uses a multi-layer perceptron (MLP) to compute the final quality scores, as shown in \cref{framework}(a). This module consists of 7 layers, \textit{i.e.}, three linear layers for dimensionality reduction, two ReLU activations for non-linearity, and two dropout layers to prevent overfitting. Before inputting to the MLP, we apply average pooling to extract the channel-wise average feature of $\tilde{\textbf{\textit{F}}}_i$, generating a channel-level global representation. We then concatenate the global representations across different semantic levels.
 The quality regression process is formulated as:
\begin{equation} 
\setlength{\abovedisplayskip}{3pt}
\setlength{\belowdisplayskip}{2pt}
\hat{Q} = \operatorname{MLP}\left(\operatorname{avg}(\tilde{\textbf{\textit{F}}}_1) \oplus\operatorname{avg}(\tilde{\textbf{\textit{F}}}_2) \oplus \cdots \oplus \operatorname{avg}(\tilde{\textbf{\textit{F}}}_n)\right), 
\end{equation} 
where $\oplus$ is the concatenation operator and the $\operatorname{avg}$ denotes the average pooling operation.

\section{Experiment}

\begin{table}[]
    \centering
    \caption{Detailed model specifications of the ESIQAnet variants, categorized as Micro (M), Tiny (T), Small (S), and Base (B). The specifications include the number of blocks at each stage, the channels used, attention heads, parameter counts (\#Param.), and floating point operations per second (FLOPs).} \vspace{-4pt}
    \label{varints}
    \vspace{-4pt}
    \setlength{\tabcolsep}{0.67em}
\scalebox{0.7}{
    \begin{tabular}{l|ccc|cc}
\toprule \text { Model } & \text { Blocks } & \text { Channels } & \text { Heads } & \#Param. & FLOPs \\
\midrule \text {  ESIQAnet-M } & {[2, 2, 8, 4]} & {[48, 96, 192, 384]} & {[2, 4, 8, 16]} & 77.57M & 7.84G \\
\text { ESIQAnet-T } & {[2, 4, 12, 4]} & {[64, 128, 256, 512]} & {[2, 4, 8, 16]} & 92.78M & 14.74G \\
\text { ESIQAnet-S } & {[3, 4, 21, 5]} & {[64, 128, 256, 512]} & {[2, 4, 8, 16]} & 104.71M & 18.90G \\
\text { ESIQAnet-B } & {[3, 4, 21, 5]} & {[96, 192, 384, 768]} & {[3, 6, 12, 24]} & 155.01M & 38.64G \\
\bottomrule
    \end{tabular}
    }
    \vspace{-17pt}
\end{table}

\begin{table*}[!t]
\centering
\vspace{-8pt}
\caption{Performance comparison of the state-of-the-art NR IQA models and the four ESIQAnet variants on our ESIQAD under three display modes. In the top-5 results for each criteria, the best result is marked in {\color[HTML]{FF0000} {\textbf{RED}}}, the second-best result is marked in {\color[HTML]{0070C0} {\textbf{BLUE}}}, and the remaining three are marked in \underline{Underlined}. }
\vspace{-7pt}
\label{1}
\setlength{\tabcolsep}{1.3em}
\scalebox{0.76}{
\begin{tabular}{l|c c c|c c c|c c c}
\toprule
Mode & \multicolumn{3}{c|}{2D} & \multicolumn{3}{c|}{3D-window} & \multicolumn{3}{c}{3D-immersive}\\
\midrule
Method &SRCC&KRCC&PLCC&SRCC&KRCC&PLCC&SRCC&KRCC&PLCC\\
\midrule
BMPRI \cite{bmpri} & 0.2803 & 0.1798 & 0.5525 & 0.1255 & 0.0825 & 0.2841 & 0.0850 & 0.0556 & 0.2780 \\ 
BPRI \cite{pri} & 0.1780 & 0.1127 & 0.2422 & 0.0925 & 0.0637 & 0.1380 & 0.0779 & 0.0535 & 0.1400 \\ 
BPRI-LSSn \cite{pri} & 0.5800 & 0.4180 & 0.6255 & 0.4287 & 0.3028 & 0.5768 & 0.3550 & 0.2472 & 0.4703 \\ 
BPRI-LSSs \cite{pri} & 0.1400 & 0.0746 & 0.3675 & 0.1036 & 0.0741 & 0.1499 & 0.1189 & 0.0823 & 0.1675 \\ 
BPRI-PSS \cite{pri} & 0.2115 & 0.1446 & 0.4519 & 0.2348 & 0.1624 & 0.4153 & 0.1923 & 0.1320 & 0.2749 \\ 
BRISQUE \cite{mittal2012no} & 0.5594 & 0.3972 & 0.6642 & 0.3156 & 0.2186 & 0.4960 & 0.2858 & 0.1943 & 0.4623 \\ 
CORNIA \cite{6247789} & 0.4073 & 0.2837 & 0.5134 & 0.1768 & 0.1188 & 0.3515 & 0.1572 & 0.1043 & 0.3555 \\ 
FISBLIM \cite{fisblim} & 0.4543 & 0.3191 & 0.5767 & 0.2800 & 0.1943 & 0.4888 & 0.2238 & 0.1521 & 0.4743 \\ 
HOSA \cite{hosa} & 0.5076 & 0.3599 & 0.6211 & 0.2713 & 0.1849 & 0.4218 & 0.2749 & 0.1864 & 0.4298 \\ 
NIQE \cite{niqe} & 0.5665 & 0.4088 & 0.6881 & 0.3081 & 0.2106 & 0.4957 & 0.3133 & 0.2137 & 0.4977 \\ 
ILNIQE \cite{ilniqe} & 0.3348 & 0.2317 & 0.4187 & 0.3118 & 0.2144 & 0.4113 & 0.2670 & 0.1827 & 0.3702 \\ 
LPSI \cite{wu2015highly} & 0.5710 & 0.4071 & 0.6959 & 0.3986 & 0.2786 & 0.5384 & 0.3396 & 0.2349 & 0.4920 \\ 
QAC \cite{QAC} & 0.4232 & 0.2867 & 0.5638 & 0.2796 & 0.1884 & 0.3882 & 0.2005 & 0.1346 & 0.3881 \\ \midrule
CNNIQA \cite{cnniqa} & 0.5824 & 0.4238 & 0.5404 & 0.5517 & 0.4001 & 0.4738 & 0.4579 & 0.3224 & 0.4050 \\ 
CNNIQA-v16 \cite{vgg} & 0.6641 & 0.4747 & 0.7031 & 0.5958 & 0.4436 & 0.6718 & 0.5041 & 0.3666 & 0.6125 \\ 
CNNIQA-v19 \cite{vgg} & 0.6883 & 0.5022 & 0.6916 & 0.6015 & 0.4470 & 0.6747 & 0.5106 & 0.3731 & 0.6007 \\ 
CNNIQA-r18 \cite{resnet} & 0.7574 & 0.5628 & 0.7006 &  0.6120 & 0.4554 & 0.6678 & 0.5127 & 0.3754 & 0.5974 \\ 
CNNIQA-r34 \cite{resnet} & 0.6882 & 0.5107 & 0.6632 & 0.6374 & 0.4839 & 0.6492 & 0.6213 & 0.4582 & 0.6952 \\ 
WaDIQaM \cite{bosse2017deep} &  0.6598 & 0.4814 & 0.6963 & 0.6493 & 0.4688 & 0.6794 & 0.5010 & 0.3539 & 0.5702 \\
HyperIQA \cite{hyperiqa} & 0.6925 & 0.5188 & 0.6870 & 0.4901 & 0.3491 & 0.4675 & 0.4013 & 0.2808 & 0.4369 \\ 
MANIQA \cite{maniqa} & \underline{0.8237} & \underline{0.6376} & \underline{0.8753} & \underline{0.7893} & \underline{0.6016} & \underline{0.8366} & \underline{0.7379} & \underline{0.5506} & \underline{0.7810}
 \\ 
TReS \cite{tres}  & 0.7400 & 0.5620 & 0.8223 & 0.6405 & 0.4696 & 0.6974 & 0.6311 & 0.4525 & 0.6030 \\  
\rowcolor{mygray} 
ESIQAnet-M & \underline{0.8334} & \underline{0.6533} & \underline{0.8683} & \underline{0.8297} & \underline{0.6476} & {\color[HTML]{0070C0} {\textbf{0.8638}}} & \underline{0.7633} & \underline{0.5771} & \underline{0.7980}\\ 
\rowcolor{mygray} 
ESIQAnet-T & \underline{0.8427} & \underline{0.6622} & \underline{0.8814} & \underline{0.8212} & \underline{0.6384} & \underline{0.8557} & \underline{0.7739} & \underline{0.5876} & \underline{0.8129}\\ 
\rowcolor{mygray} 
ESIQAnet-S &  {\color[HTML]{FF0000} {\textbf{0.8518}}} & {\color[HTML]{FF0000} {\textbf{0.6766}}} & {\color[HTML]{0070C0} {\textbf{0.8895}}} & {\color[HTML]{0070C0} {\textbf{0.8314}}} & {\color[HTML]{0070C0} {\textbf{0.6488}}} & \underline{0.8610} & {\color[HTML]{0070C0} {\textbf{0.7797}}} & {\color[HTML]{0070C0} {\textbf{0.5927}}} & {\color[HTML]{0070C0} {\textbf{0.8138}}}\\ 
\rowcolor{mygray} 
ESIQAnet-B & {\color[HTML]{0070C0} {\textbf{0.8515}}} & {\color[HTML]{0070C0} {\textbf{0.6721}}} & {\color[HTML]{FF0000} {\textbf{0.8900}}} & {\color[HTML]{FF0000} {\textbf{0.8375}}} & {\color[HTML]{FF0000} {\textbf{0.6531}}} & {\color[HTML]{FF0000}{\textbf{0.8639}}} & {\color[HTML]{FF0000} {\textbf{0.7912}}} & {\color[HTML]{FF0000} {\textbf{0.6038}}} & {\color[HTML]{FF0000} {\textbf{0.8180}}}\\ 
\bottomrule
\end{tabular}
}
\vspace{-17pt}
\end{table*}

\subsection{Experimental Setup}
\subsubsection{Compared Methods}
\label{ssec:Benchmark Models}
\vspace{-2pt}
To evaluate the performance of our proposed model, we employ 22 state-of-the-art no-reference (NR) IQA models for comparison, which are designed for traditional 2D images. The selected models can be categorized into two groups:

\begin{itemize}
\item[$\bullet$]\textbf{Hand-crafted models:} BRISQUE \cite{mittal2012no}, CORNIA \cite{6247789}, QAC \cite{QAC}, BMPRI \cite{bmpri}, NIQE \cite{niqe}, ILNIQE \cite{ilniqe}, HOSA \cite{hosa}, FISBLIM \cite{fisblim}, LPSI \cite{wu2015highly}, BPRI-PSS \cite{pri}, BPRI-LSSs \cite{pri}, BPRI-LSSn \cite{pri}, and BPRI \cite{pri}.
\item[$\bullet$]\textbf{Deep learning-based models:} CNNIQA \cite{cnniqa}, CNNIQA-r18 \cite{resnet}, CNNIQA-r34 \cite{resnet}, CNNIQA-v16 \cite{vgg}, CNNIQA-v19 \cite{vgg},  HyperIQA \cite{hyperiqa}, WaDIQaM \cite{bosse2017deep}, MANIQA \cite{maniqa}, and TReS \cite{tres}.
\end{itemize}

\vspace{-4pt}
\subsubsection{Evaluation Metrics}
\vspace{-2pt}
We evaluate the performance of these IQA models using three evaluation criteria, \textit{i.e.,} Spearman rank correlation coefficient (SRCC), Kendall’s rank correlation coefficient (KRCC), and Pearson linear correlation coefficient (PLCC). These three metrics are utilized to measure the prediction monotonicity.
Before computing these criteria, predicted scores are normalized using a five-parameter logistic function \cite{sheikh2006statistical}:
\begin{equation}
\setlength{\abovedisplayskip}{3pt}
\setlength{\belowdisplayskip}{2pt}
   \hat{y}=\beta_1\left(0.5-\frac{1}{1+e^{\beta_2\left(y-\beta_3\right)}}\right)+\beta_4 y+\beta_5, 
\end{equation}
where $\left\{\beta_i \mid i=1,2, \ldots, 5\right\}$ are parameters to be fitted, $y$ and $\hat{y}$ indicate the predicted scores and the corresponding mapped scores, respectively.

Additionally, we utilize receiver operating characteristic (ROC) analysis \cite{7498936, 7812797} as an additional evaluation method for IQA metrics. This method assesses two key aspects, \textit{i.e.}, \textit{whether two samples distinctly differ in quality and, if so, which one is superior}. 
Through \textit{``Different vs. Similar ROC Analysis"}, we determine whether various objective metrics can distinguish image pairs with and without significant qualitative differences. We classify image pairs with significant differences into pairs with positive and negative differences. 
The \textit{``Better vs. Worse ROC Analysis"} is applied to test if various objective metrics can distinguish images with positive and negative differences. The area under the ROC curve (AUC) values of the two analyses is reported in this paper, where higher values represent better performance.


\begin{table*}[t]
\vspace{-10pt}
\centering
\caption{Performance comparison of the state-of-the-art NR IQA models and the proposed ESIQAnet-B on the two subsets of our ESIQAD under three display modes. Bold entries in {\color[HTML]{FF0000} {\textbf{RED}}}, {\color[HTML]{0070C0} {\textbf{BLUE}}}, and \textbf{BLACK} are the best, second-best, and third-best performances, respectively.}
\vspace{-8pt}
\label{2}
\setlength{\tabcolsep}{1.3em}
\scalebox{0.76}{
\begin{tabular}{l|c c c|c c c|c c c}
\toprule
\multicolumn{10}{c}{Subset of Captured Images}\\ \midrule
Mode & \multicolumn{3}{c|}{2D} & \multicolumn{3}{c|}{3D-window} & \multicolumn{3}{c}{3D-immersive}\\
\midrule
Method &SRCC&KRCC&PLCC&SRCC&KRCC&PLCC&SRCC&KRCC&PLCC\\
\midrule
BMPRI \cite{bmpri} &  0.1707 & 0.1192 & 0.0737 & 0.1358 & 0.0968 & 0.2759 & 0.1224 & 0.0866 & 0.3030  \\
BPRI \cite{pri} & 0.1795 & 0.1269 & 0.1147 & 0.1914 & 0.1332 & 0.1660 & 0.1513 & 0.1052 & 0.1235  \\ 
BPRI-LSSn \cite{pri} &  0.3699 & 0.2566 & 0.6240 & 0.2822 & 0.1944 & 0.5286 & 0.2723 & 0.1856 & 0.5093  \\ 
BPRI-LSSs \cite{pri} & 0.3512 & 0.2429 & 0.4265 & 0.3001 & 0.2115 & 0.4003 & 0.2705 & 0.1897 & 0.3745  \\
BPRI-PSS \cite{pri} & 0.1385 & 0.0927 & 0.3379 & 0.1003 & 0.0673 & 0.2709 & 0.1260 & 0.0851 & 0.2656  \\
BRISQUE \cite{mittal2012no} & 0.3644 & 0.2517 & 0.6014 & 0.2624 & 0.1809 & 0.5025 & 0.2493 & 0.1672 & 0.4766  \\
CORNIA \cite{6247789} & 0.2040 & 0.1370 & 0.4603 & 0.2173 & 0.1484 & 0.3917 & 0.1692 & 0.1127 & 0.3633  \\
FISBLIM \cite{fisblim} & 0.2372 & 0.1606 & 0.6093 & 0.1988 & 0.1345 & 0.5148 & 0.1912 & 0.1277 & 0.5048  \\ 
HOSA \cite{hosa} & 0.2869 & 0.1952 & 0.4741 & 0.2154 & 0.1448 & 0.4312 & 0.2211 & 0.1482 & 0.4325  \\
NIQE \cite{niqe} & 0.3719 & 0.2568 & 0.6039 & 0.2917 & 0.1996 & 0.5162 & 0.2788 & 0.1885 & 0.5043  \\
ILNIQE \cite{ilniqe} &   0.3408 & 0.2351 & 0.4758 & 0.2801 & 0.1909 & 0.4235 & 0.2550 & 0.1734 & 0.4010  \\ 
LPSI \cite{wu2015highly} &  0.3588 & 0.2488 & 0.6274 & 0.2752 & 0.1880 & 0.5303 & 0.2655 & 0.1804 & 0.5110  \\
QAC \cite{QAC} &  0.1351 & 0.0886 & 0.3667 & 0.0917 & 0.0590 & 0.2907 & 0.0935 & 0.0604 & 0.2749  \\ \midrule
CNNIQA \cite{cnniqa} & 0.5425 & 0.3734 & 0.5296 & 0.4285 & 0.2956 & 0.4379 & 0.4006 & 0.2842 & 0.3969  \\
CNNIQA-v16 \cite{vgg} & 0.5125 & 0.3857 & 0.5498 & 0.4447 & 0.3165 & 0.5884 & 0.4341 & 0.3095 & 0.5428 \\ 
CNNIQA-v19 \cite{vgg} & 0.5580 & 0.3905 & 0.5732 & 0.4634 & 0.3297 & \textbf{0.6159} & 0.4494 & 0.3146 & 0.5687  \\ 
CNNIQA-r18 \cite{resnet} & 0.5611 & 0.3962 & 0.6006 & 0.4418 & 0.3139 & 0.5571 & 0.4115 & 0.2943 & 0.5166  \\ 
CNNIQA-r34 \cite{resnet} & 0.5522 & 0.4089 & 0.5627 & 0.4530 & 0.3399 & 0.4225 & 0.4237 & 0.3196 & 0.3852  \\ 
WaDIQaM \cite{bosse2017deep} &  0.5067 & 0.3553 & 0.5342 & 0.4595 & 0.3212 & 0.5839 & 0.3992 & 0.2751 & 0.5237  \\
HyperIQA \cite{hyperiqa} & \textbf{0.6681} & \textbf{0.4968} & 0.6341 & \textbf{0.5714} & \textbf{0.4082} & 0.5634 & 0.4462 & 0.3101 & 0.4625 \\
MANIQA \cite{maniqa} & {\color[HTML]{0070C0} {\textbf{0.7107}}} & {\color[HTML]{0070C0} {\textbf{0.5234}}} &  {\color[HTML]{0070C0} {\textbf{0.8010}}} & {\color[HTML]{0070C0} {\textbf{0.7099}}} & {\color[HTML]{0070C0} {\textbf{0.5190}}} & {\color[HTML]{0070C0} {\textbf{0.7041}}} & {\color[HTML]{0070C0} {\textbf{0.6556}}} & {\color[HTML]{0070C0} {\textbf{0.4880}}} & {\color[HTML]{FF0000} {\textbf{0.7432}}}  \\
TReS \cite{tres}  & 0.6091 & 0.4225 & \textbf{0.6618} & 0.5608 & 0.3829 & 0.5990 & \textbf{0.5424} & \textbf{0.3793} & \textbf{0.5881}  \\ 
\rowcolor{mygray} ESIQAnet-B & {\color[HTML]{FF0000} {\textbf{0.7690}}} & {\color[HTML]{FF0000} {\textbf{0.5895}}} & {\color[HTML]{FF0000} {\textbf{0.8107}}} & {\color[HTML]{FF0000} {\textbf{0.7406}}} & {\color[HTML]{FF0000} {\textbf{0.5609}}} & {\color[HTML]{FF0000} {\textbf{0.7969}}} & {\color[HTML]{FF0000} {\textbf{0.6907}}} & {\color[HTML]{FF0000} {\textbf{0.5085}}} & {\color[HTML]{0070C0}{\textbf{0.7300}}}  \\ 
\midrule
\multicolumn{10}{c}{Subset of Synthesized Images}\\ \midrule
Mode & \multicolumn{3}{c|}{2D} & \multicolumn{3}{c|}{3D-window} & \multicolumn{3}{c}{3D-immersive}\\
\midrule
Method &SRCC&KRCC&PLCC&SRCC&KRCC&PLCC&SRCC&KRCC&PLCC\\
\midrule
BMPRI \cite{bmpri} & 0.4273 & 0.2962 & 0.5067 & 0.3175 & 0.2178 & 0.4564 & 0.3412 & 0.2259 & 0.4706  \\
BPRI \cite{pri} & 0.1052 & 0.0715 & 0.3629 & 0.0562 & 0.0372 & 0.2133 & 0.0124 & 0.0081 & 0.2556  \\
BPRI-LSSn \cite{pri} & 0.4280 & 0.2929 & 0.4762 & 0.4331 & 0.3018 & 0.4232 & 0.4477 & 0.3115 & 0.4336  \\ 
BPRI-LSSs \cite{pri} & 0.1990 & 0.1313 & 0.2795 & 0.1041 & 0.0671 & 0.2249 & 0.1282 & 0.0857 & 0.2363  \\ 
BPRI-PSS \cite{pri} &  0.1011 & 0.0723 & 0.3269 & 0.0223 & 0.0182 & 0.1320 & 0.0175 & 0.0093 & 0.1774  \\ 
BRISQUE \cite{mittal2012no} &  0.2608 & 0.1677 & 0.3113 & 0.3713 & 0.2545 & 0.4232 & 0.3820 & 0.2465 & 0.4416  \\
CORNIA \cite{6247789} &  0.2130 & 0.1333 & 0.4492 & 0.1168 & 0.0804 & 0.4264 & 0.1497 & 0.1055 & 0.4318  \\
FISBLIM \cite{fisblim} & 0.0872 & 0.0541 & 0.4282 & 0.0389 & 0.0222 & 0.3712 & 0.0618 & 0.0408 & 0.3248  \\
HOSA \cite{hosa} &   0.4051 & 0.2816 & 0.4648 & 0.4112 & 0.2877 & 0.4241 & 0.4024 & 0.2731 & 0.4374  \\
NIQE \cite{niqe} & 0.4328 & 0.3164 & 0.5342 & 0.4569 & 0.3204 & 0.4822 & \textbf{0.4855} & \textbf{0.3430} & 0.5056  \\
ILNIQE \cite{ilniqe} & 0.1460 & 0.1030 & 0.2174 & 0.1935 & 0.1329 & 0.2214 & 0.1608 & 0.1127 & 0.2066  \\
LPSI \cite{wu2015highly} & 0.3952 & 0.1180 & 0.3327 & 0.3720 & 0.2400 & 0.3212 & 0.3833 & 0.2400 & 0.3263  \\
QAC \cite{QAC} & 0.2410 & 0.1604 & 0.3027 & 0.1140 & 0.0735 & 0.2116 & 0.0822 & 0.0590 & 0.2108  \\ \midrule
CNNIQA \cite{cnniqa} & 0.3286 & 0.2474 & 0.3671 & 0.2789 & 0.2368 & 0.2734 & 0.2256 & 0.1789 & 0.2269  \\ 
CNNIQA-v16 \cite{vgg} & 0.4256 & 0.3133 & 0.5685 & 0.4004 & 0.2797 & 0.5336 & 0.3936 & 0.2747 & 0.4538  \\
CNNIQA-v19 \cite{vgg} &  0.4413 & 0.2987 & 0.5732 & 0.4594 & 0.2941 & 0.5126 & 0.4278 & 0.2789 & 0.3970  \\ 
CNNIQA-r18 \cite{resnet} & 0.4429 & 0.3842 & 0.4806 & 0.3098 & 0.2737 & 0.3902 & 0.3541 & 0.2579 & 0.3669  \\ 
CNNIQA-r34 \cite{resnet} & 0.4854 & 0.3358 & 0.5030 & 0.3946 & 0.2494 & 0.4016 & 0.3675 & 0.2438 & 0.3922 \\ 
WaDIQaM \cite{bosse2017deep} &  0.4977 & 0.3615 & 0.5385 & 0.4746 & 0.3601 & 0.4716 & 0.4487 & 0.3025 & 0.4275  \\
HyperIQA \cite{hyperiqa} &  0.4624 & 0.3632 & 0.4934 & 0.4331 & 0.3158 & 0.4786 & 0.4346 & 0.2947 & 0.3660  \\
MANIQA \cite{maniqa} &  {\color[HTML]{0070C0} {\textbf{0.7081}}} & {\color[HTML]{0070C0} {\textbf{0.5323}}} & {\color[HTML]{0070C0} {\textbf{0.7380}}} & {\color[HTML]{0070C0} {\textbf{0.6976}}} & {\color[HTML]{0070C0} {\textbf{0.5006}}} & {\color[HTML]{0070C0} {\textbf{0.7819}}} & {\color[HTML]{0070C0} {\textbf{0.6537}}} & {\color[HTML]{0070C0} {\textbf{0.4670}}} & {\color[HTML]{FF0000} {\textbf{0.6687}}}  \\ 
TReS \cite{tres}  & \textbf{0.6070} & \textbf{0.4263} & \textbf{0.5795} & \textbf{0.5131} & \textbf{0.3681} & \textbf{0.6354} & 0.4738 & 0.3293 & \textbf{0.5833}  \\ 
\rowcolor{mygray} ESIQAnet-B & {\color[HTML]{FF0000} {\textbf{0.7534}}} & {\color[HTML]{FF0000} {\textbf{0.5895}}} & {\color[HTML]{FF0000} {\textbf{0.7861}}} & {\color[HTML]{FF0000} {\textbf{0.7370}}} & {\color[HTML]{FF0000} {\textbf{0.5575}}} & {\color[HTML]{FF0000} {\textbf{0.7983}}} & {\color[HTML]{FF0000} {\textbf{0.6863}}} & {\color[HTML]{FF0000} {\textbf{0.5032}}} & {\color[HTML]{0070C0} {\textbf{0.6545}}}  \\ 
\bottomrule
\end{tabular}
 } \vspace{-17pt}

\end{table*}

\begin{figure*}[ht]
\vspace{-8pt}
  \centering
  \subfloat[Different vs. Similar Analysis]{
    \begin{minipage}{0.42\linewidth}
      \centering
      \includegraphics[width=\linewidth]{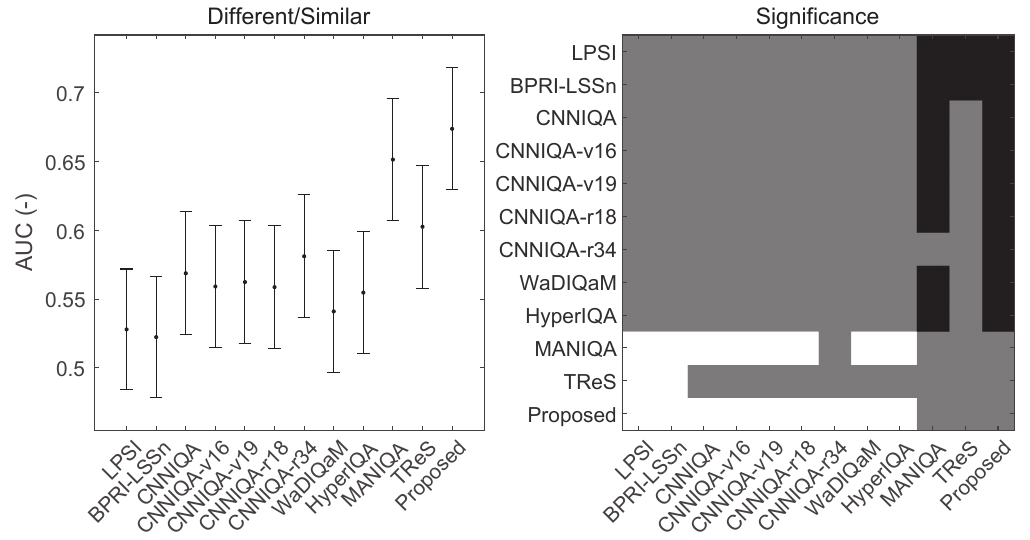}
      \hspace{2pt} 
      \includegraphics[width=\linewidth]{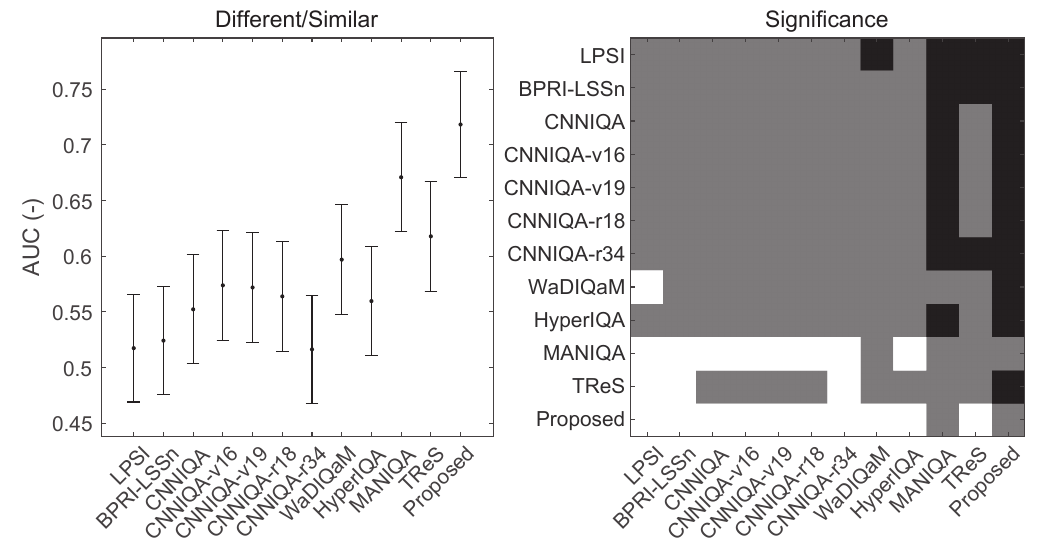}
    \end{minipage}
  } \hspace{2pt}
  \subfloat[Better vs. Worse Analysis]{
    \begin{minipage}{0.42\linewidth}
      \centering
      \includegraphics[width=\linewidth]{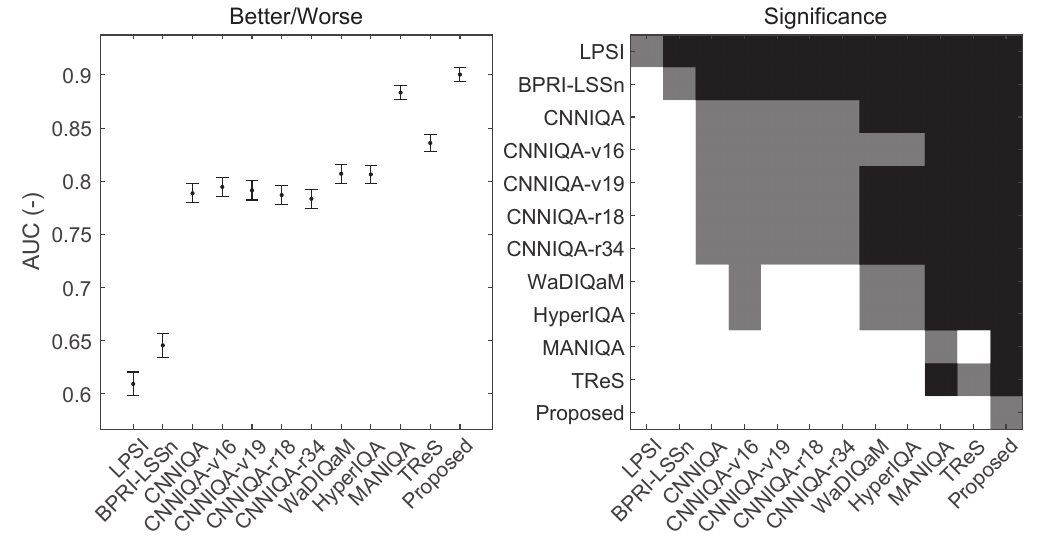}
      \hspace{2pt}
      \includegraphics[width=\linewidth]{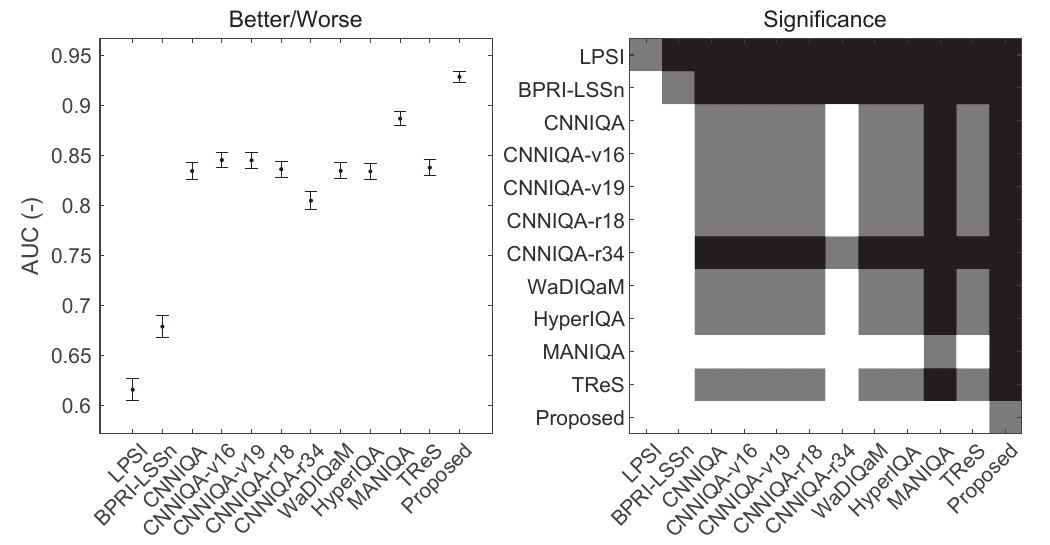}
    \end{minipage}
  }
  \vspace{-8pt}
  \caption{ROC analysis results of 11 outstanding IQA benchmark methods and the ESIQAnet on the ESIQAD in the 3D-immersive (first row) and the 3D-window (second row) display modes. 
  Note that a white/black square in the significance figures means the row metric is statistically better/worse than the column one. A gray square means the row method and the column method are statistically indistinguishable.}
  \label{ROC}
  \vspace{-5pt}
\end{figure*}

\vspace{-2pt}
\subsubsection{Implementation Details}
\vspace{-2pt}
For traditional hand-crafted benchmarks IQA models, we apply these models to directly predict the quality scores of the left and right views of an egocentric spatial image, respectively. The predicted score from the left view, which is displayed in the 2D mode, directly serves as the 2D quality score for the image. 
For both 3D-window and 3D-immersive modes, the quality score for the egocentric spatial image is calculated by averaging the scores from the left and right views.
For deep learning-based benchmarks IQA models, we train and test these models on our ESIQAD. We partition the database into training and testing sets with a ratio of 4:1. The training parameters are set the same as those in the officially released version. 
Considering that these deep learning-based IQA models are designed for traditional 2D images, we feed the left view of spatial images into the models to predict quality scores for the 2D display mode. 
Since spatial images cannot be directly input into these models, we fine-tune the architectures of these deep learning-based models to accommodate a dual-image (6 channels) input, comprising both left and right views of the spatial image. These fine-tuned models are then utilized to predict quality scores of spatial images under 3D-window and 3D-immersive display modes.

For our ESIQAnet, we extract multi-scale features after three VSSD stages with $N_1$, $N_2$, and $N_3$ VSSD blocks respectively, and after the final stage with $N_4$ MSA blocks. 
We design four architecture variants for ESIQAnet with different parameters of VSSD and MSA blocks, which are shown in \cref{varints}. When training models for different display modes, we use their corresponding MOSs as labels. Notably, for training in the 2D display mode, we use only the left view of the spatial image as input and omit the cross attention module, while other components of the model remain unchanged. 
We use the same training and test sets as the benchmark models and exclusively use the training set to train the models and obtain the optimal models that yield the lowest loss. These optimal models are then evaluated on the testing set. 


\vspace{-2pt}
\subsection{Performance Analysis}

\begin{table*}[t]
\centering
\caption{Ablation study on the attention mechanisms our ESIQAnet-B. TAB refers to the transposed attention block, CAB refers to our cross attention block, and MSAB indicates the multi-head self attention block. Best performances are indicated with \textbf{bold}.}
\vspace{-8pt}
\label{ablation}
\setlength{\tabcolsep}{1em}
\scalebox{0.76}{
\begin{tabular}{c c c c|c c c|c c c|c c c}
\toprule
\multicolumn{4}{c|}{Mode} & \multicolumn{3}{c|}{2D} & \multicolumn{3}{c|}{3D-window} & \multicolumn{3}{c}{3D-immersive}\\
\midrule
\# & CAB & TAB & MSAB &SRCC&KRCC&PLCC&SRCC&KRCC&PLCC&SRCC&KRCC&PLCC\\
\midrule
1 & & & & 0.8040 & 0.6244 & 0.8312 & 0.7898 & 0.6063 & 0.8252 & 0.7544 & 0.5665 & 0.7988  \\ 
2 & & & $\checkmark$ & 0.8221 & 0.6361 & 0.8663 & 0.8167 & 0.6100 & 0.8388 & 0.7642 & 0.5791 & 0.8039  \\
3 & & $\checkmark$ & & 0.8230 & 0.6330 & 0.8510 & 0.8023 & 0.6182 & 0.8285 & 0.7556 & 0.5698 & 0.8023  \\ 
4 & & $\checkmark$ & $\checkmark$ & \textbf{0.8515} & \textbf{0.6721} & \textbf{0.8900} & 0.8230 & 0.6384 & 0.8490 & 0.7782 & 0.5904 & 0.8128  \\
5 & $\checkmark$ & & & - & - & - & 0.8083 & 0.6250 & 0.8451 & 0.7742 & 0.5850 & 0.8045  \\
6 & $\checkmark$ & & $\checkmark$ & - & - & - & 0.8281 & 0.6403 & 0.8559 & 0.7828 & 0.5918 & 0.8137 \\
7 & $\checkmark$ & $\checkmark$ & & - & - & - & 0.8236 & 0.6428 & 0.8502 & 0.7824 & 0.5946 & 0.8063  \\
8 & $\checkmark$ & $\checkmark$ & $\checkmark$ & - & - & - & \textbf{0.8375} & \textbf{0.6531} & \textbf{0.8639} & \textbf{0.7912} & \textbf{0.6038} & \textbf{0.8180}  \\
\bottomrule
\end{tabular}
}
\vspace{-12pt}
\end{table*}

\begin{table}[!t]
\centering
\caption{Ablation study on the backbone of our ESIQAnet-B. The best performances are indicated with \textbf{bold}. The second-best performances are marked in \underline{Underlined}.}
\vspace{-6pt}
\label{ablation2}
\setlength{\tabcolsep}{0.88em}
\scalebox{0.76}{
\begin{tabular}{l |c c|c c|c c}
\toprule
Mode & \multicolumn{2}{c|}{2D} & \multicolumn{2}{c|}{3D-window} & \multicolumn{2}{c}{3D-immersive}\\
\midrule
Backbone &SRCC&PLCC&SRCC&PLCC&SRCC&PLCC\\
\midrule
ResNet-18 \cite{resnet} & 0.8162 & 0.8369 & 0.7938 & 0.8322 & 0.7315 & 0.7808   \\ 
ResNet-34 \cite{resnet} & 0.8141 & 0.8471 & 0.7907 & 0.8466 & 0.7251 & 0.7703  \\
ResNet-50 \cite{resnet} & 0.8224 & 0.8525 & 0.7986 & 0.8326 & 0.7408 & 0.7969  \\ 
ViT \cite{dosovitskiy2020image} & \underline{0.8309} & \underline{0.8601} & \underline{0.8198} & \underline{0.8527} & \underline{0.7552} & \underline{0.8031} \\
SwinT \cite{liu2021swin} & 0.8290 & 0.8522 & 0.8040 & 0.8340 & 0.7431 & 0.7913 \\
\rowcolor{mygray} VSSD & \textbf{0.8515} & \textbf{0.8900} & \textbf{0.8375} & \textbf{0.8639} & \textbf{0.7912} & \textbf{0.8180}\\ 
\bottomrule
\end{tabular}}
\vspace{-20pt}
\end{table}

\subsubsection{Evaluation on the ESIQAD}
\vspace{-2pt}
We evaluate the aforementioned 22 state-of-the-art models and our proposed model on the ESIQAD. The experimental results are demonstrated in \cref{1}. 
The results show that the four variants of ESIQAnet outperform all benchmarks across all three display modes, demonstrating excellent prediction ability and strong generalization capabilities in different display modes. Specifically, the ESIQAnet-B achieves a 2.78\% improvement in SRCC and a 1.47\% improvement in PLCC over the best benchmark model MANIQA in the 2D display mode, a 4.82\% improvement in SRCC and a 2.73\% increase in PLCC in the 3D-window mode, and a 5.33\% improvement in SRCC and a 3.70\% rise in PLCC in the 3D-immersive mode.
From \cref{1}, we observe that the performance of ESIQAnet improves with the increased number of parameters and computational complexity. More blocks enable the extraction of deeper visual semantic features, while more attention heads in the multi-head self attention mechanism help focus on various image regions simultaneously, enhancing information fusion for a more comprehensive quality assessment. For the subsequent experiments, we use the ESIQAnet-B for evaluation.

Moreover, It can be observed that most hand-crafted models show poor performance in evaluating the perceptual quality of egocentric spatial images in 2D display mode, and even perform worse for the evaluation in 3D-window and 3D-immersive modes. Compared to the hand-crafted models, deep learning-based models generally exhibit better performance across all three modes. 
For most benchmark models, the performance for 2D-display-mode quality assessment is generally better than that for the 3D-window and 3D-immersive settings. The experimental results in \cref{1} show that the ESIQAnet narrows the performance gap across these three modes, indicating its ability to well capture the stereoscopic and immersive effects of 3D scenes, which are essential for perceptual quality.
Additionally, most models exhibit slightly improved performance in predicting quality in 3D-window mode compared to 3D-immersive mode, which manifests that 3D-immersive mode presents greater challenges in predicting its QoE.

\cref{ROC} compares the AUC performance of the  ESIQAnet-B and other 11 outstanding benchmark IQA methods on the ESIQAD under 3D-window and 3D-immersive display modes, indicating that the proposed model significantly outperforms other benchmark methods on \textit{Different vs. Similar Analysis} and \textit{Better vs. Worse Analysis} criteria.

\vspace{-2pt}
\subsubsection{Evaluation on the subsets of ESIQAD}
We split the ESIQAD into the subset of 400 captured images and the subset of 100 synthesized images, evaluating the performance of benchmarks and our model on each subset separately. This approach provides additional insights into the performance of IQA models in different egocentric scenarios. Since the small size of the subset of synthesized images can lead to overfitting and negatively affect model training, we employ data augmentation methods to expand this subset to 400 samples, matching the size of the subset of captured images. The experimental results are exhibited in \cref{2}, showing the superiority of our ESIQAnet over two subsets, followed by benchmark models MANIQA, HyperIQA, and TReS, suggesting that the ESIQAnet is a robust quality metric for distinct egocentric categories under different display modes.

The experimental results indicate that the prediction performance on the subset of captured images is relatively better than on the subset of synthesized images. The captured images from the Apple Vision Pro are more realistic, featuring natural lighting, texture, and details that provide the model with rich perceptual information. In contrast, synthesized images may lack some of these natural features, making quality assessment more challenging. Additionally, the captured images align more closely with the natural scene statistics (NSS) distribution of real-world images, making them easier for the model to characterize. However, synthesized images may introduce artifacts or features that deviate from this distribution in the synthesis process, leading to reduced prediction accuracy.

\vspace{-4pt}
\subsection{Ablation Study}
\vspace{-3pt}
To validate the effect of each component of our proposed method, we conduct ablation studies on the backbone and the three attention mechanisms on ESIQAD. The results of ablation experiments are illustrated in \cref{ablation2} and \cref{ablation}.

\subsubsection{Attention Mechanisms}
\paragraph{Cross Attention Block}
The cross attention blocks in the model integrate features extracted from stereo image pairs of egocentric spatial images, utilizing their disparity to capture interactive features. To evaluate the effectiveness of cross attention blocks, we conducted ablation experiments by replacing the cross attention aggregation with simple averaging of features from the left and right views. As shown in \cref{ablation}, the cross attention block proves to be essential, enhancing the model's performance when combined with the other two modules.
\vspace{-2pt}
\paragraph{Transposed Attention Block}
In the ESIQAnet, the transposed attention block is employed to facilitate global interactions across channels, enabling more comprehensive information exchange throughout the model. By introducing transposed attention blocks, we can observe a clear performance gain in SRCC and PLCC, highlighting the critical role of capturing inter-channel dependencies.
\vspace{-2pt}
\paragraph{Multi-head Self Attention Block}
The multi-head self attention block is utilized to capture spatial dependencies and complex local relationships within features. The results in \cref{ablation} show that the multi-head self attention block improves the model's ability to learn detailed and robust feature representations, resulting in enhanced performance.
\vspace{-3pt}
\subsubsection{Architecture Backbone} \vspace{-2pt}
In the ESIQAnet, we employ multi-stage VSSD blocks that are pre-trained on ImageNet-1K \cite{deng2009imagenet} to extract content-aware spatial features from binocular views of egocentric spatial images. We compare our method with other pre-trained networks for spatial feature extraction, including ResNet-18 \cite{resnet}, ResNet-34 \cite{resnet}, ResNet-50 \cite{resnet}, Vision Transformer (ViT) \cite{dosovitskiy2020image}, and Swin Transformer (SwinT) \cite{liu2021swin}, all pre-trained on ImageNet. 
For a fair comparison, we freeze the parameters of these models during feature extraction, which is the same as the ESIQAnet model. The results illustrated in \cref{ablation2} suggest that the backbone of multi-stage VSSD blocks used in ESIQAnet surpasses the other visual feature extraction networks.
\vspace{-5pt}
\section{Conclusion}
\vspace{-2pt}
In this work, we conduct a comprehensive study for egocentric image quality assessment.
We establish the first egocentric spatial images database, termed ESIQAD, comprising 500 egocentric spatial images along with their quality ratings under three common display modes, \textit{i.e.,} 2D, 3D-window, and 3D-immersive modes. 
Based on our ESIQAD, we analyze the human visual perception of these images across various display modes. Furthermore, we propose a mamba2-based multi-stage feature fusion model ESIQAnet to predict the perceptual quality of egocentric spatial images across three display modes. 
The results from our extensive experiments indicate that the ESIQAnet outperforms 22 state-of-the-art benchmark IQA models on the ESIQAD and its subsets, manifesting that our ESIQAnet is a robust quality metric for different egocentric categories under different display modes.


\textbf{Applications.} The ESIQAnet addresses a critical need in IQA research, as traditional 2D metrics fail to capture the unique features of egocentric spatial images, \textit{e.g.,} binocular disparity, immersion effects, and first-person perspectives, which are key aspects essential for AR/VR applications. The primary goal of ESIQAnet is to monitor and optimize QoE by accurately predicting the perceptual quality of egocentric spatial images. It can serve as a versatile tool across the media pipeline, functioning as a performance evaluation metric for codecs to optimize compression while preserving user experience. It can also guide content adaptation for various display devices and assist in real-time quality monitoring for live streaming, ensuring consistent image quality and enhancing QoE in immersive applications like gaming, virtual training, and telepresence, \textit{etc.}

\textbf{Limitations and Future Work.} The database currently only contains egocentric spatial images captured using Apple Vision Pro and iPhone devices. Although these devices were chosen for their advanced imaging capabilities, the lack of diversity in capture devices limits the generalizability of the database. Our future works will extend the database by including egocentric spatial images captured with alternative devices, such as Meta Quest 3, Varjo XR-3, or other head-mounted displays, to improve its diversity and applicability across a broader range of devices. 
Furthermore, the potential impact of motion sickness in egocentric content is an important aspect to explore, especially for egocentric spatial videos viewed in immersive VR environments. It is important to study how motion sickness may affect the QoE and integrate this factor into the evaluation of egocentric spatial videos, providing a more comprehensive understanding of user experience in immersive media in future work.


\vspace{-5pt}
\acknowledgments{%
\vspace{-3pt}
    This work was supported in part by the National Key R\&D Program of China under Grant 2021YFE0206700, in part by the National Natural Science Foundation of China under Grants 62401365, 62271312, 62225112, 62132006, and in part by the Shanghai Pujiang Program under Grant 22PJ1407400.
}

\vspace{-6pt}
\section*{Human Research Statement}
\vspace{-3pt}
Our subjective experiments only require human participants to watch everyday materials under commercially available non-hazardous equipment for a limited time and provide their feedback. Consequently, this subjective experiment is safe and has no negative impact on the human participants.
Moreover, this work does not contain any personally identifiable information or sensitive data. All data is anonymized and does not involve any infringement on personal privacy. The research primarily focuses on technology development, model testing, and algorithm optimization, with no direct impact on human subjects. The research adheres to all relevant ethical standards and legal regulations, ensuring that data usage and handling comply with ethical requirements.
Therefore, IRB approval is not required in this subjective experiment. 
In addition, we have obtained the consent of human subjects in the study by signing an informed consent form before the subjective experiment. All participants were fully informed about the content of the experiments, their voluntary involvement, and the use of their feedback.

\bibliographystyle{abbrv-doi-hyperref}

\bibliography{main}








\end{document}